%% file: main.tex
\documentclass[twoside]{article}

\usepackage{multirow}
\usepackage{graphicx}
\usepackage{caption}
\usepackage{subcaption}
\usepackage{amsfonts}
\usepackage{dirtytalk}
\usepackage{float}  % Add the float package for better float placement
\usepackage[accepted]{aistats2024}
\usepackage[round]{natbib}

\begin{document}



\twocolumn[

\aistatstitle{Unveiling the Limits of Learned Local Search
Heuristics: Are You the Mightiest of the Meek?}

\aistatsauthor{ Ankur Nath \And Alan Kuhnle }

\aistatsaddress{ Department of Computer Science and Engineering, Texas A\&M University} ]

% \begin{abstract}
% \end{abstract}

\input{Sections/abstract}
\input{Sections/Introduction}
\input{Sections/related_work}
\input{Sections/evalution}

\input{Sections/summary_and_outlook}

\bibliography{reference.bib}
\newpage
\onecolumn
\section*{SUPPLEMENTARY MATERIALS}

\subsection*{Reproducibility}

We use the publicly available implementation of ECO-DQN\footnote{Code available at: https://github.com/tomdbar/eco-dqn} and GNNSAT\footnote{Code available at: https://github.com/emreyolcu/sat} as our code base. To ensure a fair comparison, we employ the pretrained models from the original papers. However, for S2V-DQN applied to ER graphs with $|V|=60$, we conduct training from scratch due to the inability to load the pretrained model with the hyperparameters provided in the original paper. We provide all our experimental results, code, and data at this link\footnote{Code available at: https://tinyurl.com/52ykxtaj}.

For training skewed graphs for the Max-Cut problem, we adopt the hyperparameters outlined in the original paper and modify the dataset while keeping other settings intact.

\paragraph{Experimental Setup} We run all experiments on two NVIDIA RTX A4000 (16GB memory) GPUs with Intel Xeon(R) w5-2445 CPUs at 3.10 GHz and 64 GB of memory

\section*{Additional Tables and Plots}

In this section, you can find the tables and plots omitted in the main paper due to space constraints.

\subsection*{Intra-episode Behavior}

\begin{figure}[h]

     \centering
     
     \begin{subfigure}[H]{0.3\textwidth}
         
         \centering
         \includegraphics[width=\textwidth]{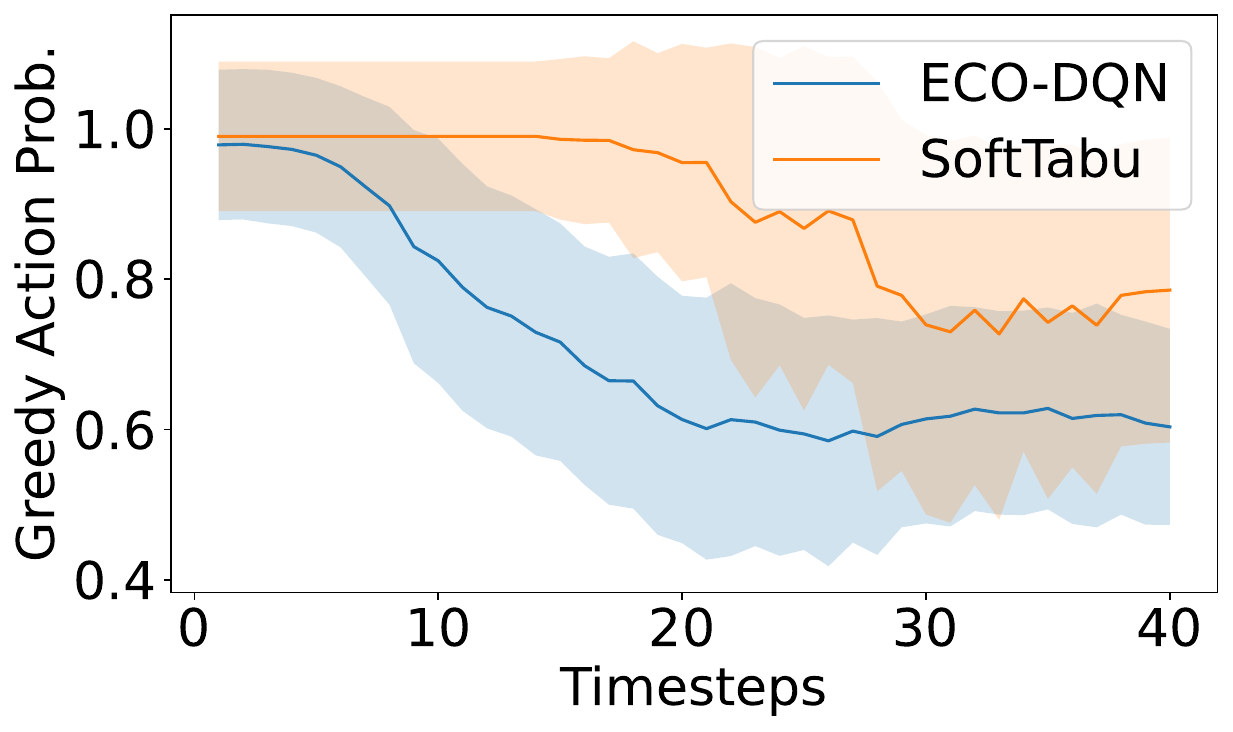}
         \caption{ER20 agent}
         % \label{fig:y equals x}
     \end{subfigure}
     % \hfill
     \begin{subfigure}[H]{0.3\textwidth}
        
         \centering
         \includegraphics[width=\textwidth]{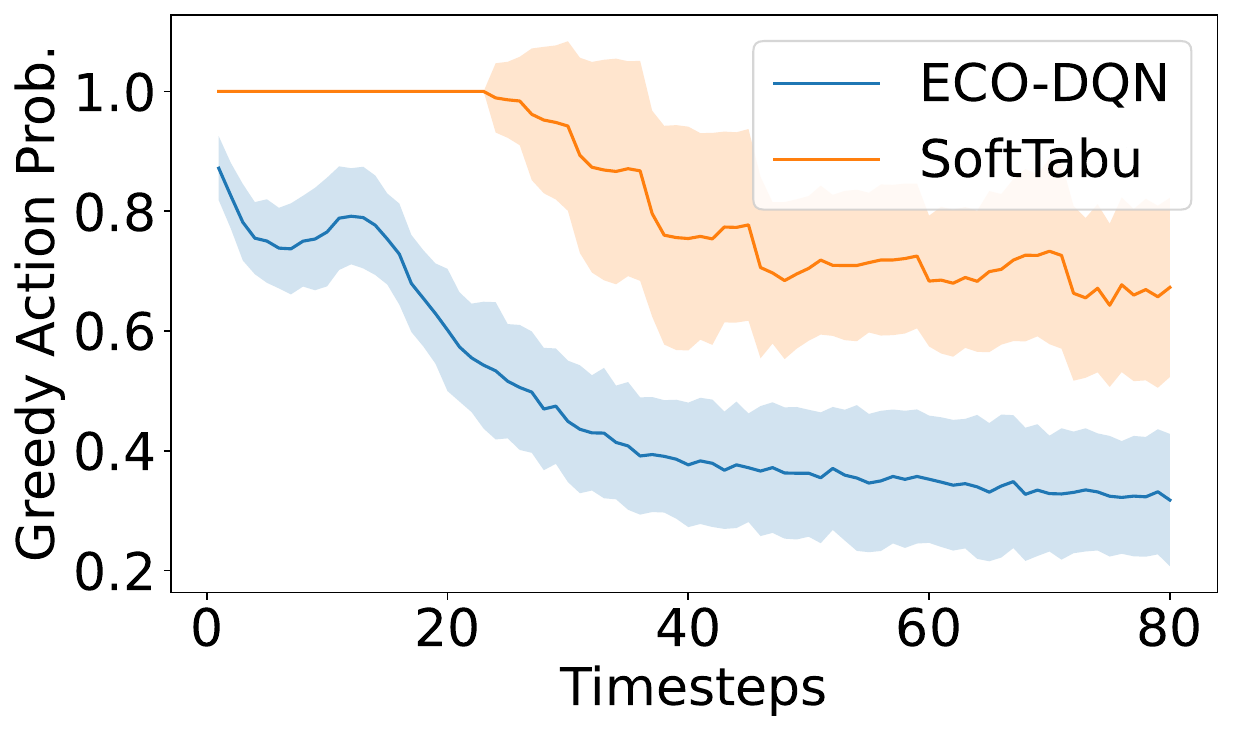}
         % \caption{$y=3\sin x$}
         \caption{ER40 agent}
         % \label{fig:three sin x}
     \end{subfigure}
      \begin{subfigure}[H]{0.3\textwidth}
         
         \centering
         \includegraphics[width=\textwidth]{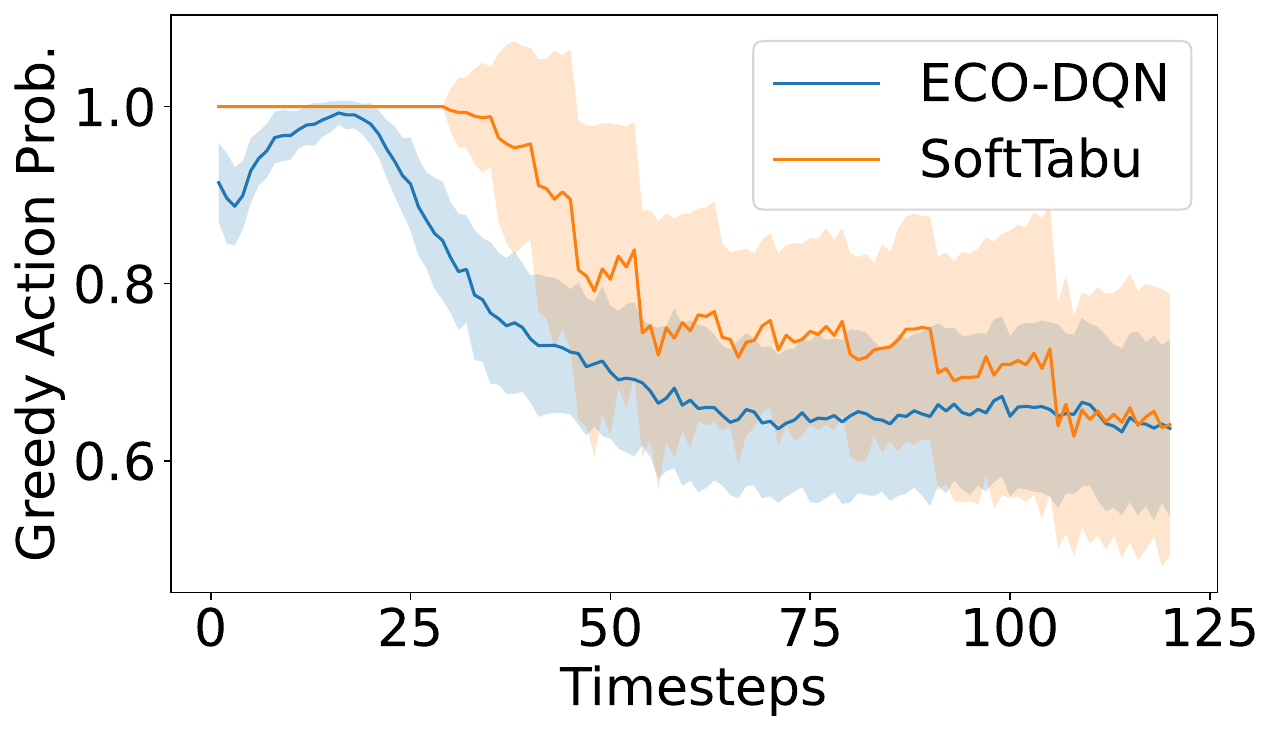}
         \caption{ER60 agent}
         % \label{fig:y equals x}
     \end{subfigure}
     \hfill

     \begin{subfigure}[H]{0.3\textwidth}
        
         \centering
         \includegraphics[width=\textwidth]{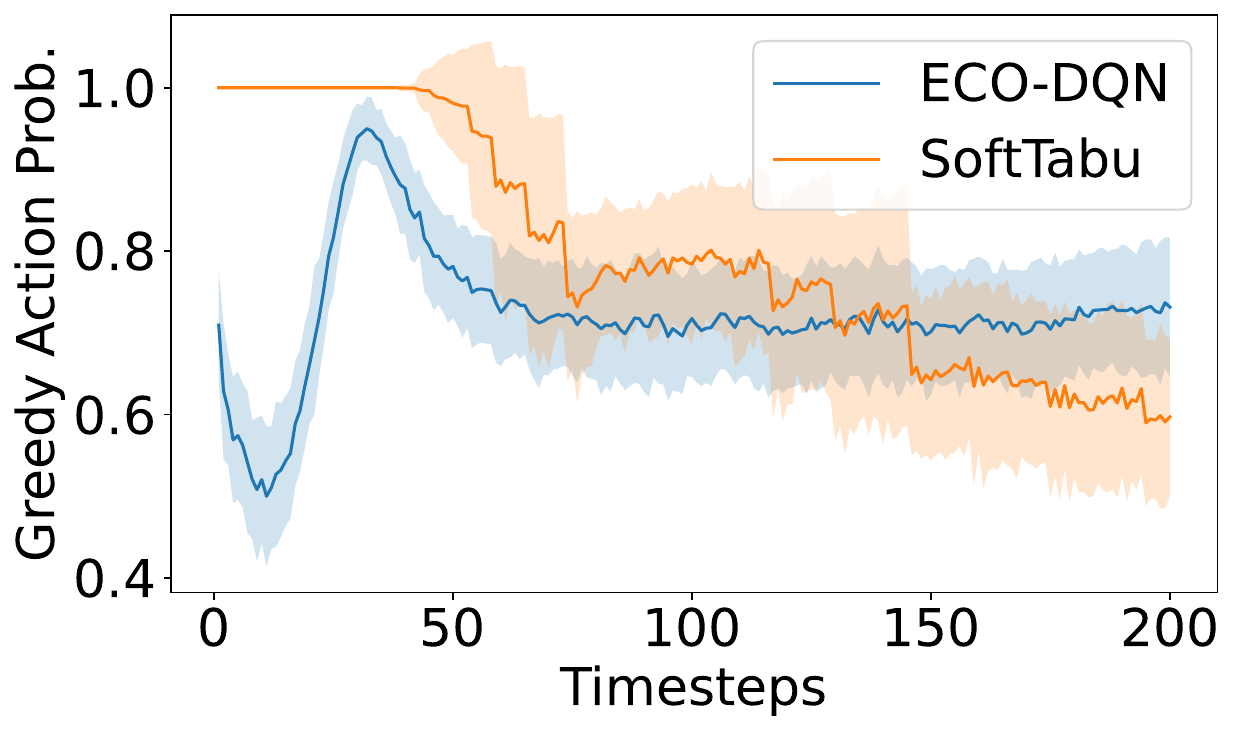}
         % \caption{$y=3\sin x$}
         \caption{ER100 agent}
         % \label{fig:three sin x}
     \end{subfigure}
      % \hfill
     \begin{subfigure}[H]{0.3\textwidth}
        
         \centering
         \includegraphics[width=\textwidth]{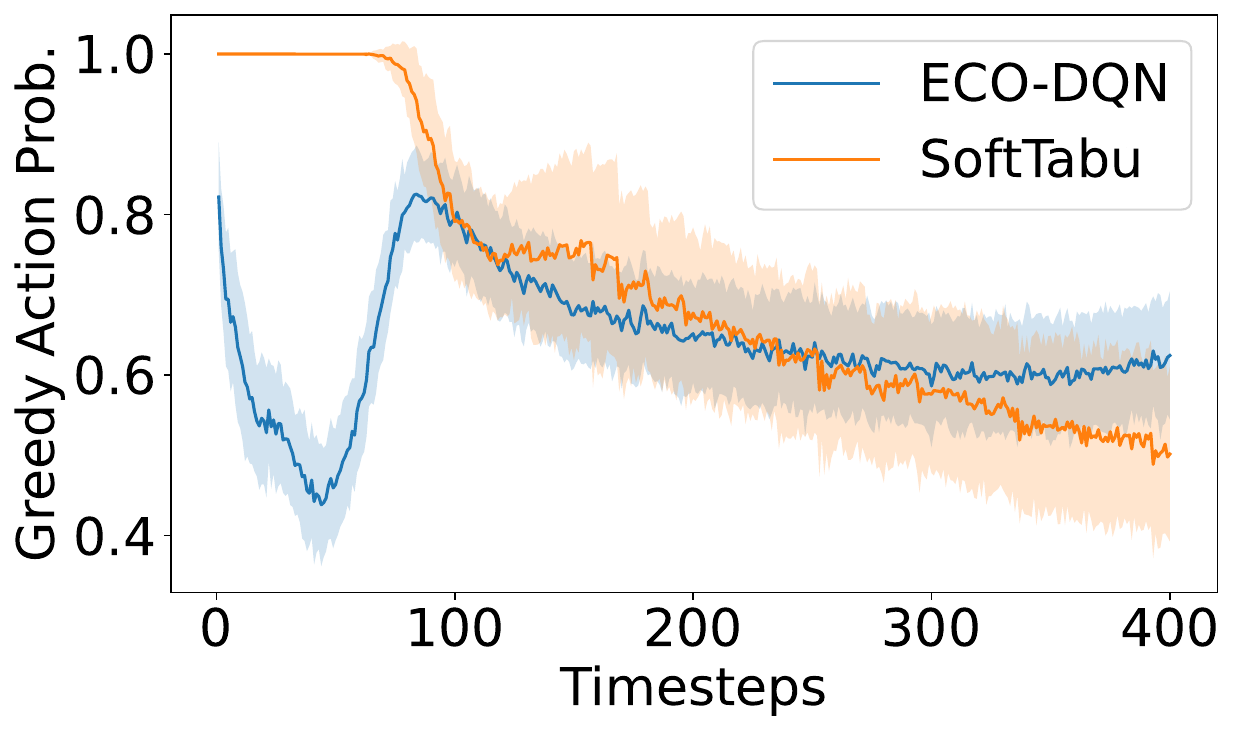}
         % \caption{$y=3\sin x$}
         \caption{ER200 agent}
         % \label{fig:three sin x}
     \end{subfigure}
     \hfill
     \caption{Intra-episode behavior of ECO-DQN and SoftTabu agents averaged across all 100
instances from the validation set of ER graphs with  from $|V | =
20$ to $200$.}
\label{fig:Intra-episode behaviorER}
     % \vspace{1em}
        
\end{figure}

\begin{figure}[h]

     \centering
     
     \begin{subfigure}[H]{0.3\textwidth}
         
         \centering
         \includegraphics[width=\textwidth]{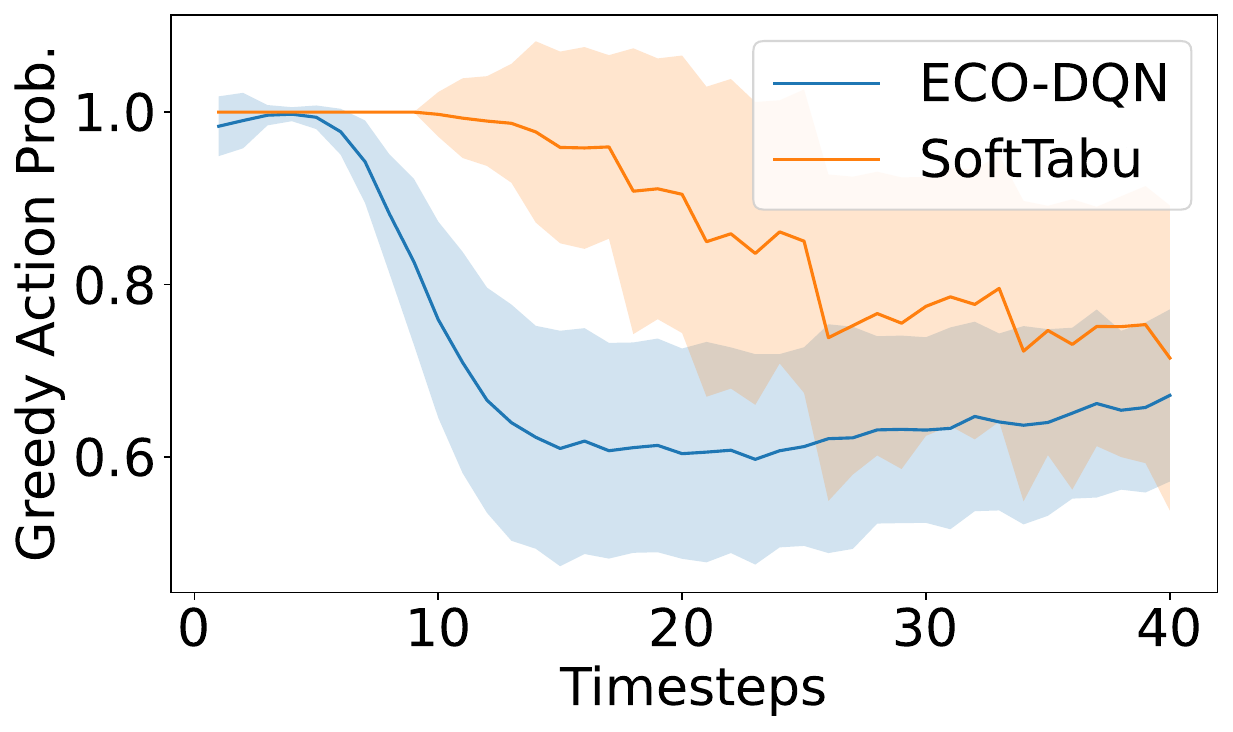}
         \caption{BA20 agent}
         % \label{fig:y equals x}
     \end{subfigure}
     % \hfill
     \begin{subfigure}[H]{0.3\textwidth}
        
         \centering
         \includegraphics[width=\textwidth]{Images/intra-episode/Intra_episodeBA20.pdf}
         % \caption{$y=3\sin x$}
         \caption{BA40 agent}
         % \label{fig:three sin x}
     \end{subfigure}
      \begin{subfigure}[H]{0.3\textwidth}
         
         \centering
         \includegraphics[width=\textwidth]{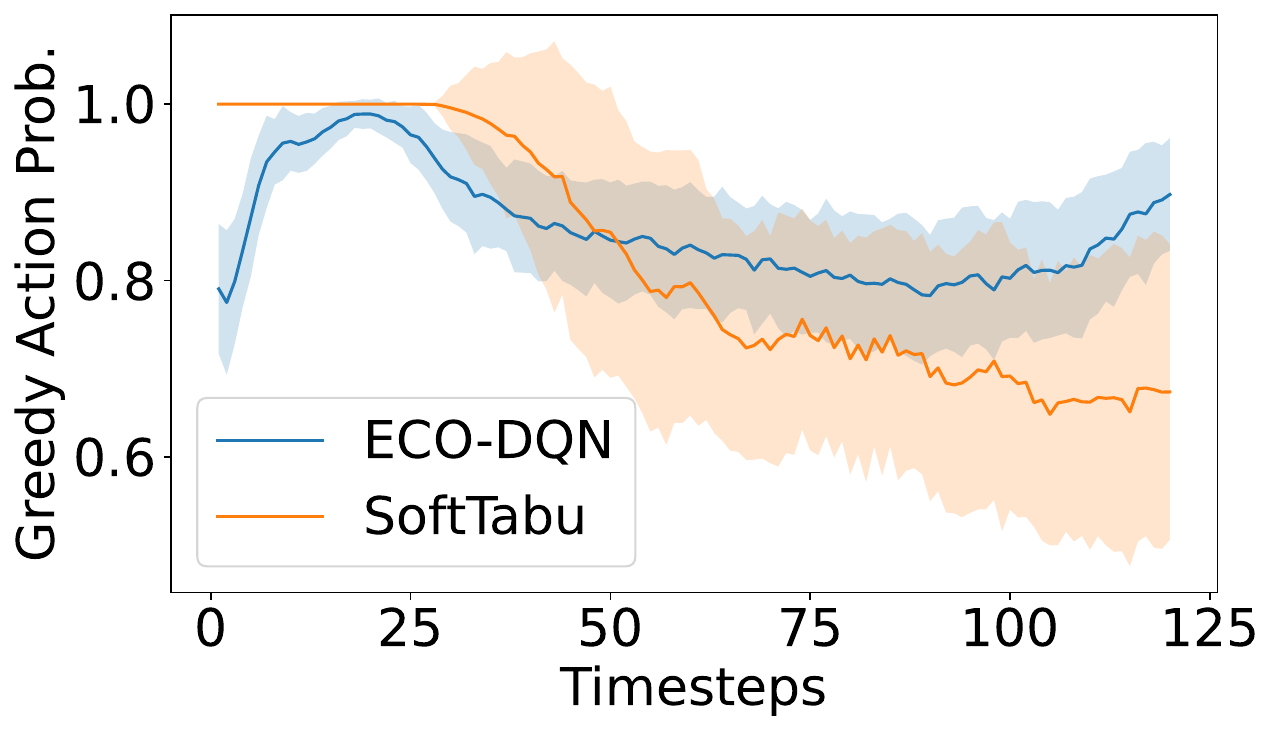}
         \caption{BA60 agent}
         % \label{fig:y equals x}
     \end{subfigure}
     \hfill

     \begin{subfigure}[H]{0.3\textwidth}
        
         \centering
         \includegraphics[width=\textwidth]{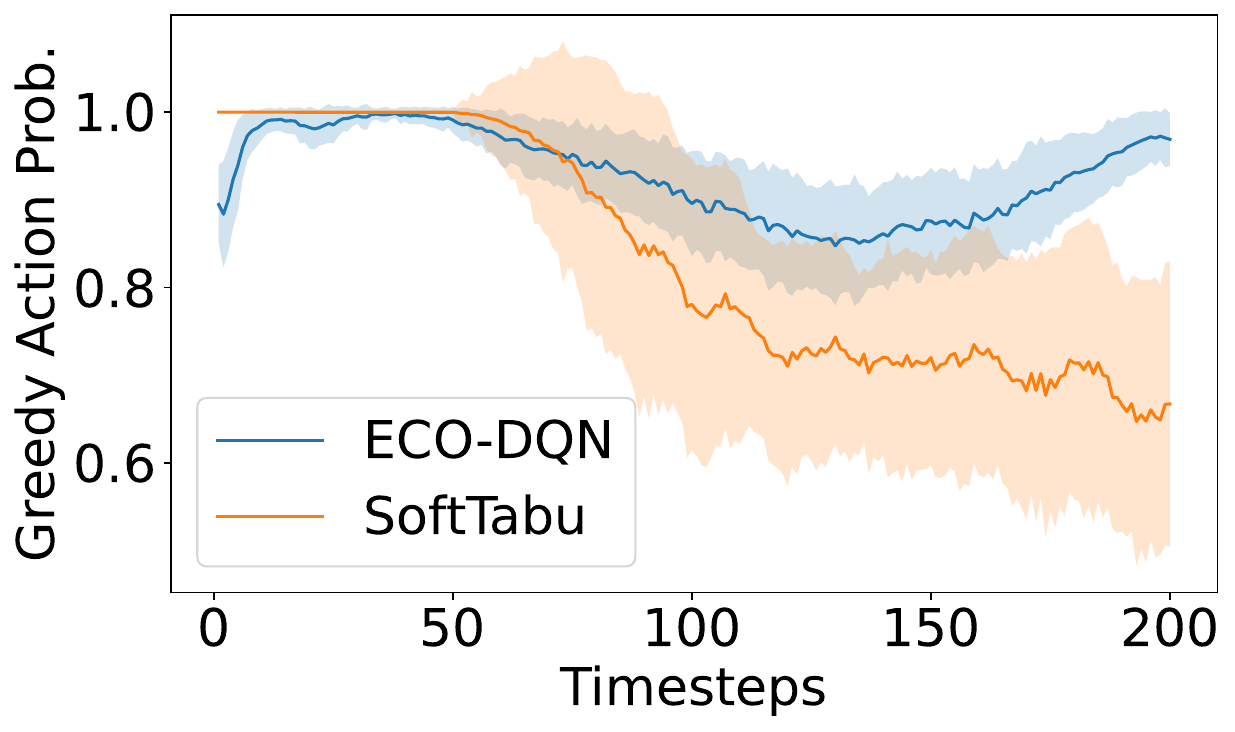}
         % \caption{$y=3\sin x$}
         \caption{BA100 agent}
         % \label{fig:three sin x}
     \end{subfigure}
      % \hfill
     \begin{subfigure}[H]{0.3\textwidth}
        
         \centering
         \includegraphics[width=\textwidth]{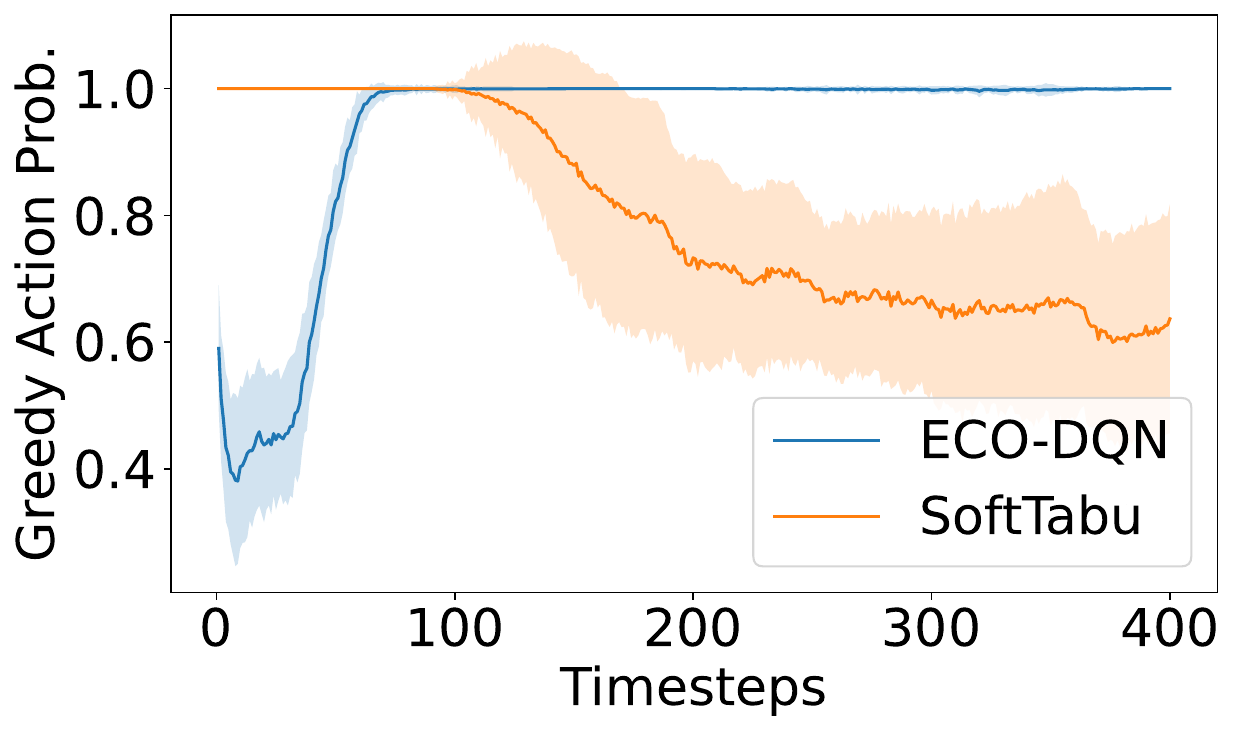}
         % \caption{$y=3\sin x$}
         \caption{BA200 agent}
         % \label{fig:three sin x}
     \end{subfigure}
     \hfill
     \caption{Intra-episode behavior of ECO-DQN and SoftTabu agents averaged across all 100
instances from the validation set of BA graphs with  from $|V | =
20$ to $200$. }
\label{fig:Intra-episode behaviorBA}
     % \vspace{1em}
        
\end{figure}

Figure \ref{fig:Intra-episode behaviorER} and \ref{fig:Intra-episode behaviorBA} present an  analysis of the intra-episode behavior of ECO-DQN and SoftTabu agents across a diverse range of distributions.

\subsection*{Generalisation on Small Instances}
Understanding the generalization of agents is crucial, as it determines their practical utility in real-world applications where the data distribution may vary or evolve over time.Table \ref{fig:ERTable} and \ref{fig:BATable} provide detailed analysis of how well these agents can adapt and perform in scenarios that go beyond their training data.

\begin{table}[H]
\centering
\caption{Generalisation of agents trained on ER graphs of size $|V |=40 $ to unseen graph sizes and structures.}
\label{fig:ERTable}
% \label{tab:my-table}

\begin{tabular}{ccccc}
\hline
      & Tabu  & S2V-DQN & ECO-DQN & SoftTabu                        \\ \hline
ER60  & 1.0 & 0.97 & 1.0 & 1.0  \\
ER100 & 1.0 & 0.96 & 1.0 & 1.0 \\
ER200 & 1.0 & 0.95 & 1.0 & 1.0   \\
ER500 & 0.99 & 0.92 & 0.99 & 0.99                          \\ \hline
BA40  & 1.0 & 0.97 & 1.0 & 1.0  \\
BA60  & 1.0 & 0.97 & 1.0 & 1.0 \\
BA100 & 1.0 & 0.94 & 1.0 & 1.0  \\
BA200 & 0.98 & 0.86 & 0.98 & 0.98   \\
BA500 & 0.96 & 0.74 & 0.97 & 0.98\\ \hline
\end{tabular}
\end{table}

\begin{table}[H]
\centering
\caption{Generalisation of agents trained on BA graphs of size $|V |=40 $ to unseen graph sizes and structures.}
\label{fig:BATable}

\begin{tabular}{ccccc}
\hline
      & Tabu  & S2V-DQN & ECO-DQN & SoftTabu                        \\ \hline
ER40  & 1.0 & 0.97 & 1.0 & 1.0  \\
ER60  & 1.0 & 0.95 & 1.0 & 1.0 \\
ER100 & 1.0 & 0.94 & 1.0 & 1.0  \\
ER200 & 1.0 & 0.93 & 0.99 & 1.0  \\
ER500 & 1.0 & 0.9 & 0.98 & 0.99                           \\ \hline
BA60  & 1.0 & 0.96 & 1.0 & 1.0   \\
BA100 & 1.0 & 0.94 & 1.0 & 1.0   \\
BA200 & 0.98 & 0.81 & 0.98 & 0.98   \\
BA500 & 0.97 & 0.5 & 0.99 & 0.99 \\ \hline
\end{tabular}
\end{table}

\newpage
\subsection*{Generalisation on Hard Instances and Real World Datasets}

Table \ref{hard-benenchmarks} presents a performance analysis of agents that have been trained on skew graphs, specifically those with  $|V| = 200$. The primary focus is on how these agents perform when faced with established benchmark tasks, with the best-performing results being emphasized and displayed in bold for clarity.
\begin{table}[H]
\centering
\caption{ Average performance of agents trained on skew graphs of size $|V |=200$ on known benchmarks (best in bold).}
\label{hard-benenchmarks}
\begin{tabular}{ccccccc}
\hline
Dataset & Type     & $|V|$ & Tabu  & S2V-DQN & ECO-DQN & SoftTabu \\ \hline
Physics & Regular  & 125   & \textbf{1.0}   & 0.908   & \textbf{1.0}     & \textbf{1.0 }     \\
G1-10   & ER       & 800   & \textbf{0.991} & 0.927   & 0.98    & 0.983    \\
G11-G13 & Torodial & 800   & 0.965 & 0.9     & 0.968   & \textbf{0.988}   \\
G14-G21 & Skew     & 800   & 0.967 & 0.887   & 0.955   & \textbf{0.975}    \\
G22-31  & ER       & 2000  & 0.978 & 0.93    & 0.969   & \textbf{0.981}    \\
G32-34  & Torodial & 2000  & 0.937 & 0.904   & 0.965   & \textbf{0.983}    \\
G35-42  & Skew     & 2000  & 0.95  & 0.663   & 0.945   & \textbf{0.968}   \\ \hline
\end{tabular}
\end{table}

% \section{MISSING PROOFS}

% The supplementary materials may contain detailed proofs of the results that are missing in the main paper.

% \subsection{Proof of Lemma 3}

% \textit{In this section, we present the detailed proof of Lemma 3 and then [ ... ]}

% \section{ADDITIONAL EXPERIMENTS}

% If you have additional experimental results, you may include them in the supplementary materials.

% \subsection{The Effect of Regularization Parameter}

% \textit{Our algorithm depends on the regularization parameter $\lambda$. Figure 1 below illustrates the effect of this parameter on the performance of our algorithm. As we can see, [ ... ]}

% \vfill

\subsection*{Distributions of Actions on Hard Instances}
\begin{figure}[h]

     \centering
     
     \begin{subfigure}[H]{0.3\textwidth}
         
         \centering
         \includegraphics[width=\textwidth]{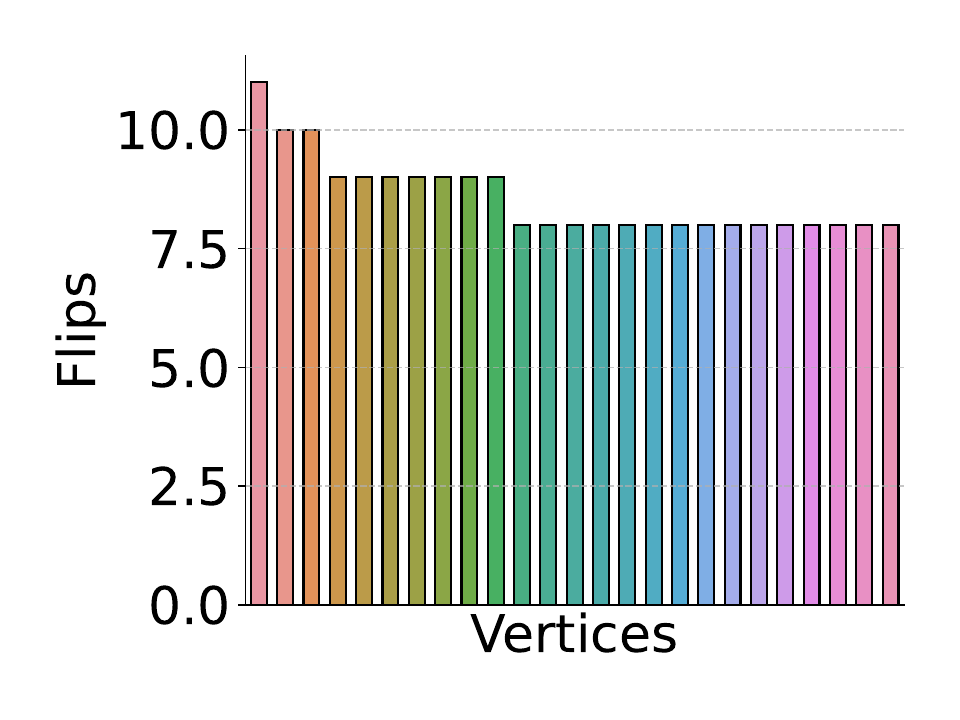}
         \caption{ER (train), ER (test)}
         % \label{fig:y equals x}
     \end{subfigure}
     % \hfill
     \begin{subfigure}[H]{0.3\textwidth}
        
         \centering
         \includegraphics[width=\textwidth]{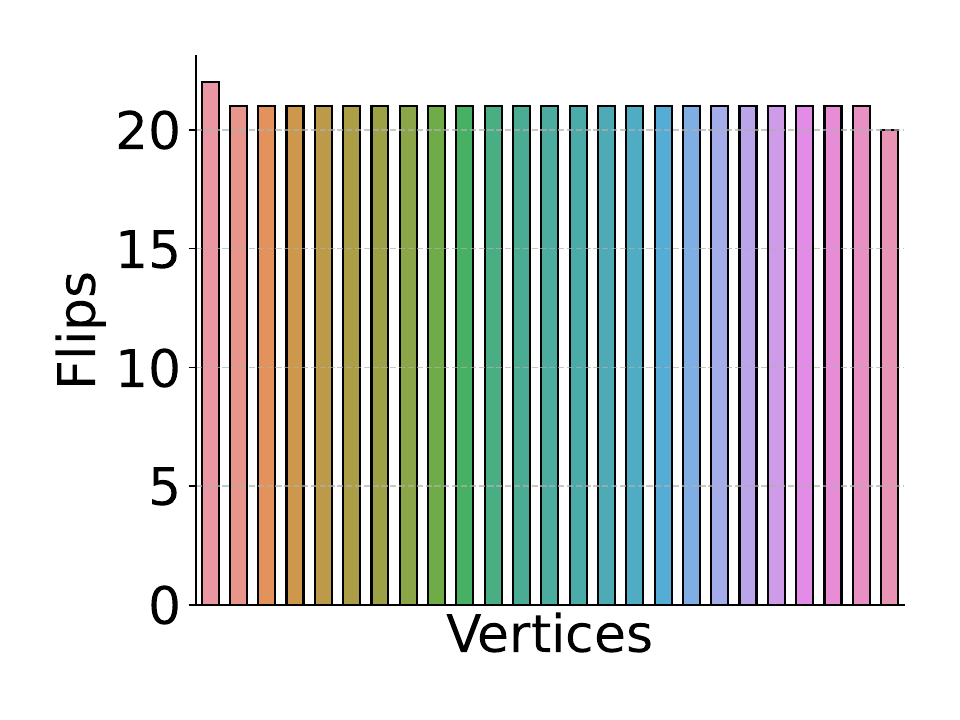}
         % \caption{$y=3\sin x$}
         \caption{ER (train), Torodial (test)}
         % \label{fig:three sin x}
     \end{subfigure}
      \begin{subfigure}[H]{0.3\textwidth}
         
         \centering
         \includegraphics[width=\textwidth]{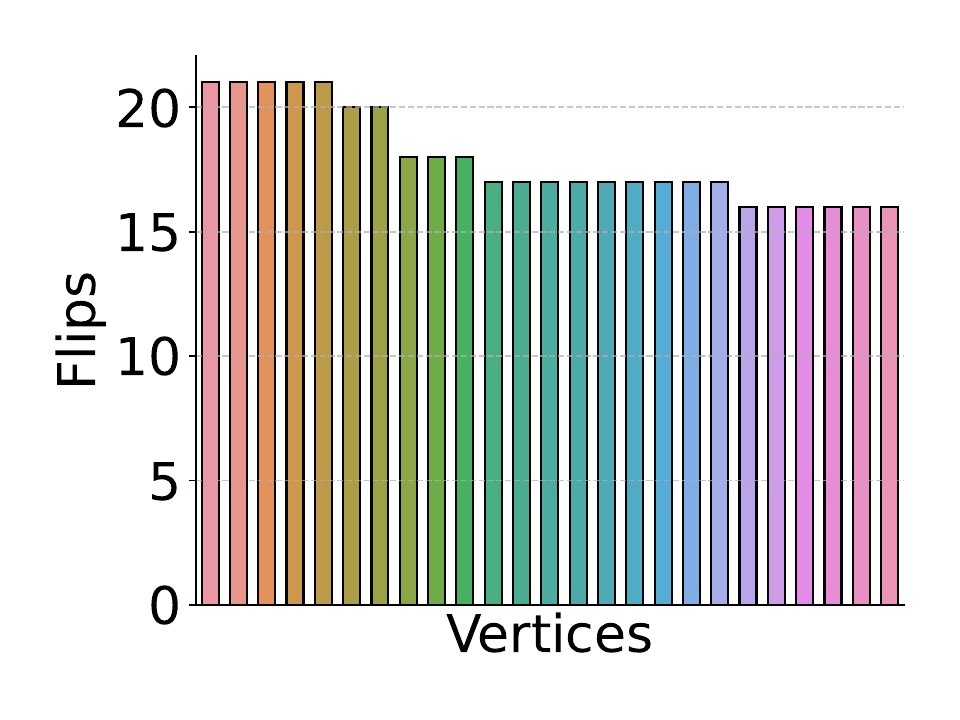}
         \caption{ER (train), Skew (test)}
         % \label{fig:y equals x}
     \end{subfigure}
     \hfill
     \caption{Distribution of flips (number of times a
vertex state is changed during an episode) of SoftTabu agents in descending order on a random graph
from three distributions with $|V | = 2000 $ from GSET
Dataset, Trained on ER with $|V | =
200$. We limit the number of vertices to $25 $.}
\label{fig:Intra-episode behavior}
     % \vspace{1em}
        
\end{figure}

\begin{figure}[h]

     \centering
     
     \begin{subfigure}[H]{0.3\textwidth}
         
         \centering
         \includegraphics[width=\textwidth]{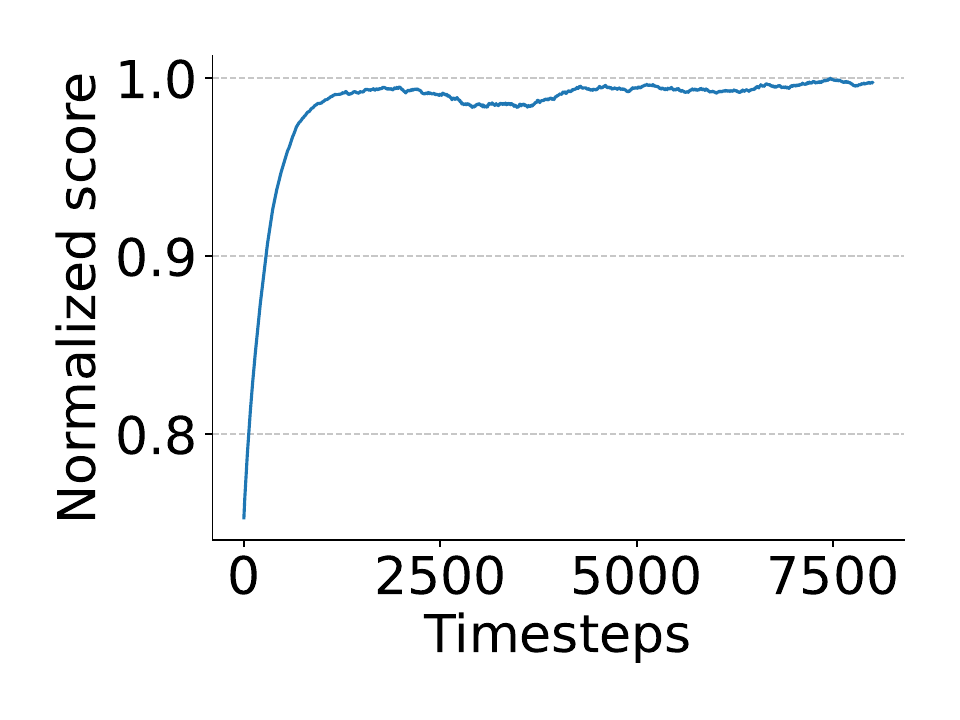}
         \caption{ER (train), ER (test)}
         % \label{fig:y equals x}
     \end{subfigure}
     % \hfill
     \begin{subfigure}[H]{0.3\textwidth}
        
         \centering
         \includegraphics[width=\textwidth]{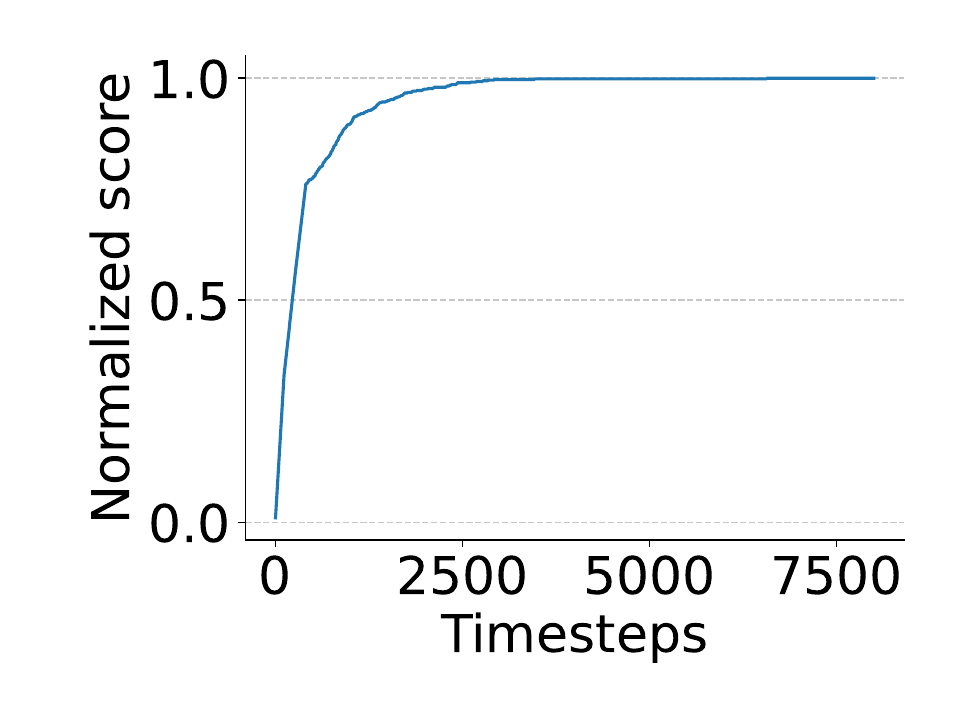}
         % \caption{$y=3\sin x$}
         \caption{ER (train), Torodial (test)}
         % \label{fig:three sin x}
     \end{subfigure}
      \begin{subfigure}[H]{0.3\textwidth}
         
         \centering
         \includegraphics[width=\textwidth]{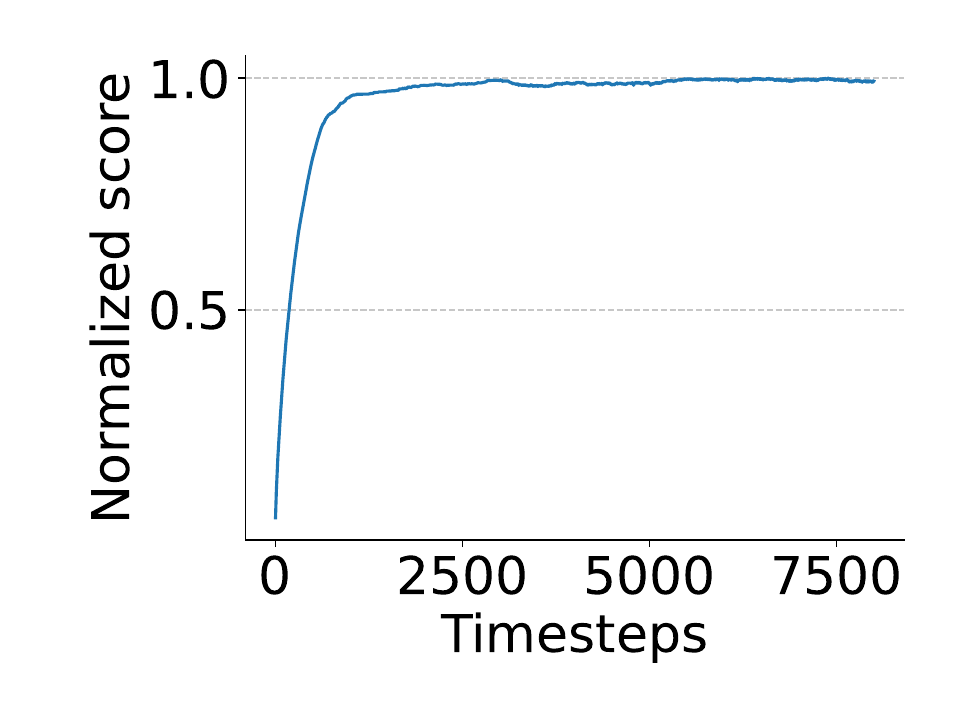}
         \caption{ER (train), Skew (test)}
         % \label{fig:y equals x}
     \end{subfigure}
     \hfill
     \caption{ Score of SoftTabu agents on a random graph
from three distributions with $|V | = 2000 $ from GSET
Dataset, Trained on ER with $|V | =
200$.}
\label{fig:Intra-episode behavior}
        
\end{figure}

\subsection*{Evalution of SAT}
In Table \ref{sat}, we compare the performance of learned heuristics alongside the WalkSAT algorithm. For WalkSAT, we have set the value of the parameter $p$ to 0.5, following the experimental setup used in GNNSAT. Additionally, we have fine-tuned the algorithm to determine the optimal $p$ value.

\begin{table}[H]
\centering
\caption{Performance of the learned heuristics and WalkSAT. For WalkSAT, we set the value of $p = 0.5 $ following the experimental setup of GNNSAT and also tune the algorithm for the optimal value. In each cell, there are three metrics (from top to bottom): the ratio of the average number of steps, the median number of steps, and the percentage solved. *Values as reported in GNNSAT for reference.\newline}
\label{sat}

\begin{tabular}{cccccccc}
\hline
Distribution                          & SoftTabu &  & GNNSAT* &  & WalkSAT (p=0.5) &  & WalkSAT (optimal p) \\ \hline
\multirow{3}{*}{\(\ \text{rand\textsubscript{3}}(50,213)\)}  & 273      &  & 367     &  & 433             &  & 408                 \\
                                      & 185      &  & 273     &  & 449             &  & 362                 \\
                                      & $96\%$   &  & $84\%$  &  & $78\%$          &  & $82\%$              \\ \hline
\multirow{3}{*}{\(\text{clique\textsubscript{3}}(20,0.05)\)} & 126      &  & 116     &  & 239             &  & 202                 \\
                                      & 42       &  & 57      &  & 185             &  & 130                 \\
                                      & $100\%$  &  & $100\%$ &  & $100\%$         &  & $100\%$             \\ \hline
\multirow{3}{*}{$\text{color\textsubscript{5}}(20,0.5)$}     & 190      &  & 342     &  & 442             &  & 352                 \\
                                      & 77       &  & 223     &  & 434             &  & 272                 \\
                                      & $99\%$   &  & $88\%$  &  & $80\%$          &  & $89\%$              \\ \hline
\multirow{3}{*}{\(\text{domset\textsubscript{4}}(12,0.2)\)} & 76       &  & 205     &  & 184             &  & 144                 \\
                                      & 42       &  & 121     &  & 119             &  & 89                  \\
                                      & $100\%$  &  & $100\%$ &  & $100\%$         &  & $100\%$             \\ \hline
\end{tabular}
\end{table}

\end{document}

%% file: Sections/abstract.tex
\begin{abstract}
In recent years, combining neural networks with local search heuristics has become popular in the field of combinatorial optimization. Despite its considerable computational demands, this approach has exhibited promising outcomes with minimal manual engineering. However, we have identified three critical limitations in the empirical evaluation of these integration attempts. Firstly, instances with moderate complexity and weak baselines pose a challenge in accurately evaluating the effectiveness of learning-based approaches. Secondly, the absence of an ablation study makes it difficult to quantify and attribute improvements accurately to the deep learning architecture. Lastly, the generalization of learned heuristics across diverse distributions remains underexplored. In this study, we conduct a comprehensive investigation into these identified limitations. Surprisingly, we demonstrate that a simple learned heuristic based on Tabu Search surpasses state-of-the-art (SOTA) learned heuristics in terms of performance and generalizability. Our findings challenge prevailing assumptions and open up exciting avenues for future research and innovation in combinatorial optimization.
\end{abstract}

%% file: Sections/introduction.tex
\section{INTRODUCTION}
Designing effective heuristics or approximation algorithms for NP-hard combinatorial optimization (CO) problems is a challenging task, often requiring domain knowledge and extensive trial-and-error.
Thus, the idea of automating this demanding and tedious design process through machine learning to learn algorithms that exploit the inherent structure of a problem has gained significant interest among researchers \citep{bello2016neural,khalil2017learning,bengio2021machine,dong2021new}. Particularly, a considerable portion of these works \citep{khalil2017learning,barrett2020exploratory,yolcu2019learning} have concentrated on employing Graph Neural Networks (GNNs) for CO problems. Despite the computational demands, these GNN-based approaches have demonstrated competitive performance compared to SOTA heuristics tailored to specific problems.

While these potential advancements offer great promise, some concerns remain. The superior performance of the learned heuristics can be attributed to the selection of specific instances and baselines. Specifically, if the baselines are weak, the learned heuristics can easily outperform them. Without hard instances and proper baseline selection, the learned heuristics can easily show comparable performance with the SOTA heuristics, and this can lead to an overestimation of the actual capabilities of the learned heuristics. Moreover, comparisons with SOTA heuristics are occasionally left out due to scalability challenges with deep learning architectures.

A subset of learning-based approaches \citep{khalil2017learning,yolcu2019learning,barrett2020exploratory,barrett2022learning} incorporate the functionality or behavior of traditional heuristics, potentially offering improved or enhanced performance by integrating heuristics principles with machine learning components. If a comprehensive comparison with the integrated heuristics and an ablation study of the deep learning architecture are lacking, it becomes challenging to determine the specific contribution of the deep learning architecture. Consequently, if the integrated heuristics are robust, the learned heuristics can seamlessly outperform the baseline heuristics, while the deep learning architecture plays a little role.

Another great achievement of learned heuristics \citep{khalil2017learning,barrett2020exploratory,toenshoff2021graph}, initially trained on smaller and specific instances from a particular distribution, exhibits impressive performance when tested on larger instances from different distributions. This achievement stands as a significant accomplishment, aligning with the core objective of employing learning-based approaches: to mitigate the necessity for extensive customization and domain-specific knowledge often required by heuristics. Although classical heuristics with hyperparameters may also encounter challenges if they are fine-tuned for a specific distribution, they may also generalize across different distributions. The primary inquiry revolves around whether the learned heuristics indeed demonstrate superior generalizability compared to classical heuristics. A thorough and insightful comparison of learned heuristics against classical heuristics provides valuable insights into the generalizability of learned heuristics.

Learned heuristics often lack theoretical guarantees, making empirical evaluation the sole method to comprehend the strengths and limitations of proposed methodologies. We believe that several fundamental yet pivotal questions remain unexplored in the empirical evaluation of these works. While these inquiries are pertinent to all types of learned heuristics, we ask and answer them by analyzing highly cited and recent peer-reviewed publications focused on learning local search heuristics. Our goal is not to provide exhaustive benchmarks for the CO problems discussed in our work but rather to assist future researchers in evaluating their research.

Concretely, we ask and answer the following questions:
\begin{enumerate}
\item \textit {Can learned heuristics be over-parameterized?} Absolutely. Replacing the GNN in ECO-DQN with linear regression and utilizing a subset of features from ECO-DQN \citep{barrett2020exploratory} that links to Tabu Search \citep{glover1990tabu}, we introduce a pruned version of ECO-DQN termed SoftTabu. Our study demonstrates that SoftTabu showcases superior performance and generalizability in comparison to ECO-DQN for the NP-hard Maximum-Cut (Max-Cut) problem.
\item \textit{Can baseline bias be attributed to the superior performance of learned heuristics?} Yes, we demonstrate that SoftTabu, a vanilla learned heuristic, can outperform S2V-DQN \citep{khalil2017learning} for the Max-Cut problem and GNNSAT \citep{yolcu2019learning} for Boolean Satisfiability (SAT).
\item \textit{Can learned heuristics demonstrate superior generalizability owing to instance selection bias?} Yes, we demonstrate that ECO-DQN shows poor generalization on harder instances and easily gets trapped in the search space.
\end{enumerate}

%% file: Sections/related_work.tex
\section{RELATED WORK}

As stated before, we will limit our discussion to local search heuristics. For future reference, we will specifically refer to local search heuristics when discussing heuristics.
 
% In this section, we formally define local search heuristics: let $\mathcal{S}$ be the search space representing all possible solutions given an optimization problem, and $f:\mathcal{S}\rightarrow \mathbb{R}$ be the objective function that needs to be optimized. A local search heuristic starts with an initial solution $s_0 \in \mathcal{S}$ and a neighborhood function $N(s)$ that generates a set of neighboring solutions for a given solution $s$. The algorithm iteratively explores neighboring solutions based on a defined mechanism, typically moving to a neighbor that improves the objective function value. This process can be represented as:

% \begin{equation*}
% s_{i+1}=\text{argmax \textsubscript{$s' \in N(s_i)$}} f(s')
% \end{equation*}
% The algorithm terminates when a stopping criterion is satisfied, which could be reaching a local optimum, a predefined number of iterations, or other specified conditions.
\subsection{Classical Heuristics}

Local search heuristics are applied to a wide range of challenging CO problems across various domains, including computer science (especially artificial intelligence), mathematics, operations research, engineering, and bioinformatics. Some examples of local search heuristics are Simulated Annealing \citep{kirkpatrick1983optimization}, Tabu search \citep{glover1990tabu}, Extremal optimization \citep{boettcher2001extremal}, Hybrid GRASP heuristics \citep{festa2009hybrid}, and Breakout Local search \citep{benlic2013breakout}. However, the performance of these algorithms relies on the specific problem at hand, and achieving optimal results often demands domain-specific knowledge and a significant amount of manual engineering. 

% \subsection{Learning local search heuristics}

\subsection{Generalized Learned Heuristics}
In this subsection, we explore works that can solve diverse CO problems without polynomial time reduction. In their influential work, \cite{khalil2017learning} introduced Structure-to-Vector GNN trained with Deep Q-Networks (DQN) to address various CO problems. The resulting S2V-DQN algorithm showcased strong performance across diverse CO problems by effectively generalizing across graphs of varying sizes and topologies. Building on the work of \cite{khalil2017learning} , \cite{manchanda2019learning} initially trained an embedding Graph Convolution Network (GCN) in a supervised manner. Subsequently, they trained a Q-network using reinforcement learning (RL) to predict action values for each vertex. Their method employed GCN embeddings to prune nodes unlikely to be part of the solution set, significantly enhancing scalability compared to S2V-DQN. However, it is not applicable to problems where nodes cannot be pruned, including fundamental CO problems like the Max-Cut problem.

\cite{barrett2020exploratory} proposed ECO-DQN, a SOTA RL algorithm for Max-Cut. ECO-DQN improves the initial solution by navigating between the local optimal points. However, ECO-DQN uses a costly GNN at each decision step, resulting in worse scalability than S2V-DQN. To address scalability challenges in ECO-DQN, \cite{barrett2022learning} proposed ECORD. This approach limited the costly GNN to a pre-processing step and introduced a recurrent unit for fast-action selection.

\subsection{Learned Heuristics for SAT}
In the domain of Boolean Satisfiability, the application of machine learning to solving SAT problems is not a new idea \citep{battiti1997reactive,haim2009restart,flint2012perceptron,grozea2014can,ganesh2009avatarsat,liang2016learning}. In recent years, there has been a trend in integrating GNNs with SAT solvers\citep{yolcu2019learning,lederman2019learning,kurin2020can,jaszczur2020neural}, aiming to improve search heuristics by leveraging predictions from GNNs. 

%% file: Sections/evalution.tex
\section{EVALUATION for MAX-CUT}

\subsection{Problem Formulation}
In this subsection, we formally define the Max-Cut problem as follows: Given an undirected graph $G(V, E)$, where $V$ represents the set of vertices, $E$ denotes the set of edges and weights $w(u,v)$ on the edges $(u,v)\in E$, the goal is to find a subset of nodes $S \subseteq V$ that maximizes the objective function, $f(S)= \sum_{u \in S, v \in V\setminus S} w(u, v)$. More than half of the $21$ NP-complete problems enumerated in \cite{Karp1972} and many real-world applications \citep{perdomo2012finding,elsokkary2017financial,venturelli2019reverse} can be reduced to the Max-Cut problem.
We analyze S2V-DQN and ECO-DQN for the Max-Cut problem in both its weighted and unweighted variants.

\subsection{Datasets for Max-Cut}

In this subsection, we briefly discuss datasets included in our analysis. Our datasets as well as the experimental code can be found at this link \footnote{Code Repository: https://tinyurl.com/52ykxtaj}.

\paragraph{GSET}
Stanford GSET dataset \citep{ye2003gset} is extensively used to benchmark SOTA heuristics \citep{benlic2013breakout,leleu2019destabilization} for Max-Cut. The dataset comprises three types of weighted and unweighted random graphs: Erd{\H{o}}s-R{\'e}nyi graphs \citep{erdHos1960evolution} with uniform edge probabilities, skew graphs with decaying connectivity, and regular toroidal graphs.

\paragraph{Synthetic graphs}
We incorporate the dataset published by \cite{barrett2020exploratory}, featuring Erd{\H{o}}s-R{\'e}nyi \citep{erdHos1960evolution} and Barab{\'a}si-Albert \citep{albert2002statistical} graphs (referred to as ER and BA, respectively). These graphs involve edge weights \(w_{ij} \in \{0,\pm1 \}\) and include up to 500 vertices. Various optimization approaches \citep{CPLEX,leleu2019destabilization,tiunov2019annealing} are applied to each graph, and the best solution found by any of these approaches is considered the optimum solution.
\paragraph{Physics}
We investigate a real-world dataset known as the \say{Physics} dataset, comprising ten graphs with $125$ vertices representing Ising models. In this dataset, each vertex has six connections, and the edge weights are defined as \(w_{ij} \in \{0,\pm1 \}\). These graphs served as a generalization benchmark in \cite{khalil2017learning}.

\subsection{Learned Heuristics}
In this subsection, we will discuss the two learned heuristics to be analyzed and explain why we believe it is important to reevaluate these learned heuristics.

\paragraph{S2V-DQN} \cite{khalil2017learning} demonstrated that S2V-DQN outperforms standard greedy, CPLEX, and semidefinite programming \citep{goemans1995improved} for the Max-Cut problem. Despite the well-known poor performance of the standard greedy approach for this problem \citep{fujishige2005submodular}, the standard greedy approach was the second-best competitor to their approach.

\paragraph{ECO-DQN} \cite{barrett2020exploratory} introduced a RL framework that allows reversible actions and provided seven handcrafted observations per node to represent the state space. Here, we want to highlight two observations closely related to the Tabu Search: \romannumeral 1) immediate change of objective value (marginal gain) if vertex state is changed (a vertex is added to or removed from the solution set) and \romannumeral 2) steps since the vertex state was last changed to prevent short looping trajectories. Empirically, ECO-DQN demonstrated superior performance compared to S2V-DQN while demonstrating competitive performance with SOTA heuristics \citep {tiunov2019annealing,leleu2019destabilization}. ECO-DQN allows both greedy and non-greedy actions to perform an in-depth local search, striking a balance between exploration and exploitation. We refer to the original paper for more details on the algorithm.

Despite the comprehensive comparison with SOTA heuristics, we observed an absence of a direct comparison with its simple heuristic counterpart, Tabu Search.
Incorporating such a comparison can help understand whether ECO-DQN indeed learns a more evolved version of TS or if its superior performance might be attributed to its integration with a local search heuristic.

\subsection{Baselines}
In this subsection, we will provide a brief overview of baseline heuristics for comparison.

\paragraph{Maxcut Approximation (MCA)}This is a simple greedy algorithm that starts with a random solution and iteratively improves the solution by choosing the optimal move at each step until no further improvement is possible. This differs from the standard greedy approach, which starts with an empty solution set and does not allow reversible actions.

\paragraph{Tabu Search (TS)}\cite{glover1990tabu} proposed a local search optimization algorithm that starts with a random solution and explores the best neighboring solutions. It involves a parameter known as \textit{tabu tenure} (denoted as $\gamma$), acting as short-term memory, restricting the repetition of the same actions for $\gamma$ number of timesteps. This prevents revisiting recent actions and allows TS to escape from local minima and explore the search space. Although there are various improved versions of this algorithm mentioned in the original paper \cite{glover1990tabu}, we chose the standard Tabu search as our baseline.

\subsection{SoftTabu (Pruned ECO-DQN)}

 We propose a simple RL framework based on TS. Within this framework, we replace the GNN component of ECO-DQN with linear regression. Additionally, we exclusively utilize a specific subset of node features from ECO-DQN that are tied to TS. This substitution enables us to conduct an empirical evaluation of the role of GNN in learning heuristics.

Given an undirected graph $G(V,E)$ and a solution $S$ at time step $t$, we formally define the state, action, and reward within our RL framework as follows:

\paragraph{State} We define the state of the environment denoted as $s^{(t)}$ by providing two observations per vertex. These observations comprise: \romannumeral 1) the marginal gain resulting if the vertex state is changed; and \romannumeral 2) the number of steps since the vertex state is changed. Notably, we intentionally omit all other features utilized in ECO-DQN.

\paragraph{Action} An action, $a^{(t)}$, involves selecting a vertex and subsequently changing its state.
\paragraph{Reward} The reward is defined as $r(s^{(t)})=\text{max}(\frac{f(s^{(t)})-f(s^{*})}{|V|},0)$, where $f(s^{*})$ stands for the best solution found in the episode so far. Additionally, whenever the agent reaches a local minima (where no action immediately increases the objective value), previously unseen in the episode, a small reward of value $1/|V|$ is provided. Our definition of reward is similar to ECO-DQN.
\paragraph{Transition}Transition is deterministic here and corresponds to changing the state of the vertex $v \in V$ that was selected as the last action.

We use Q-learning to learn a deterministic policy with a discount factor of $\gamma = 0.95$. Once trained, an approximation of the optimal policy $\pi$ can be obtained simply by acting greedily with respect to the predicted Q-values: $\pi(s^{(t)};\theta) = \text{argmax\textsubscript{$a^{(t)}$}}Q(s^{(t)},a^{(t)};\theta)$.
We want to remark that combining machine learning with TS is not a new idea \citep{battiti1994reactive,battiti1997reactive}. Our approach, SoftTabu, simply serves as a learned heuristic baseline that strongly follows TS.

\subsection{Performance Benchmarking on Small Instances}

We evaluate the algorithms using the average approximation ratio as a performance metric. Given an instance, the approximation ratio for an algorithm (agent) is the ratio of the objective value of the best solution found by the agent to the objective value of the best-known solution for the instance. Unless specifically mentioned, for each graph $G(V,E)$, we run each stochastic reversible agent for 50 randomly initialized episodes with $2|V|$ timesteps per episode and select the best outcome from these runs. We would like to stress that we run and present all empirical evaluations of ECO-DQN to ensure a fair comparison with it and are able to reproduce its performance on the synthetic datasets \citep{barrett2020exploratory} at the level reported in the original work. Due to scalability issues with GNNs, we refrain from training agents on distributions with 500 vertices.

\begin{table*}[t]

\begin{center}
\caption{Average approximation ratios of 100 validation graphs (where SoftTabu matches or outperforms ECO-DQN in bold).} \label{same-dist}

\begin{tabular}{ccccccc}
\hline
      & \multicolumn{2}{c}{Heuristics} &  & \multicolumn{3}{c}{Learned Heuristics} \\ \cline{2-3} \cline{5-7} 
      & MCA            & Tabu          &  & S2V-DQN     & ECO-DQN    & SoftTabu    \\ \hline
ER20  & 1.0            & 1.0           &  & 0.968       & 0.99       & \textbf{0.99}       \\
ER40  & 0.998          & 1.0           &  & 0.98        & 1.0        & \textbf{1.0}        \\
ER60  & 0.994          & 1.0           &  & 0.969& 1.0        & \textbf{1.0}        \\
ER100 & 0.978          & 1.0           &  & 0.924       & 1.0        & \textbf{1.0}         \\
ER200 & 0.961          & 1.0           &  & 0.952       & 1.0        & 0.999       \\ \hline
BA20  & 1.0            & 1.0           &  & 0.973       & 1.0        & \textbf{1.0}         \\
BA40  & 0.999          & 1.0           &  & 0.961       & 1.0        & \textbf{1.0}         \\
BA60  & 0.989          & 1.0           &  & 0.939       & 1.0        & \textbf{1.0}        \\
BA100 & 0.965          & 1.0           &  & 0.952       & 1.0        & \textbf{1.0}        \\
BA200 & 0.914          & 0.983         &  & 0.926       & 0.979      & \textbf{0.984}      \\ \hline
\end{tabular}
\end{center}
\end{table*}

In Table \ref{same-dist}, we present the performance of agents trained and tested on the same distribution in synthetic datasets \citep{barrett2020exploratory} up to 200 vertices. The results underscore the noteworthy observation that MCA, a relatively simple heuristic with no hyperparameters, frequently outperforms S2V-DQN for the Max-Cut problem. Looking ahead, if future research on learned heuristics utilizes S2V-DQN as a benchmark and outperforms S2V-DQN, it raises two distinct possibilities. Firstly, it could signify genuine progress in enhancing performance for solving CO problems. Alternatively, it might suggest that all these learned heuristics are weak, and none of them truly excel.

This leads to our initial concern: comparing a weak heuristic against another weak heuristic does not provide meaningful insights into their performance. Benchmarking against weak heuristics may set a low standard, giving a false sense of accomplishment if the evaluated heuristic performs better. Proper baseline selection will ensure a better grasp of the extent of generalizability and the effectiveness of learned heuristics in tackling diverse CO problems.

\begin{figure}[ht]

     \centering
     
     \begin{subfigure}[b]{0.23\textwidth}
         
         \centering
         \includegraphics[width=\textwidth]{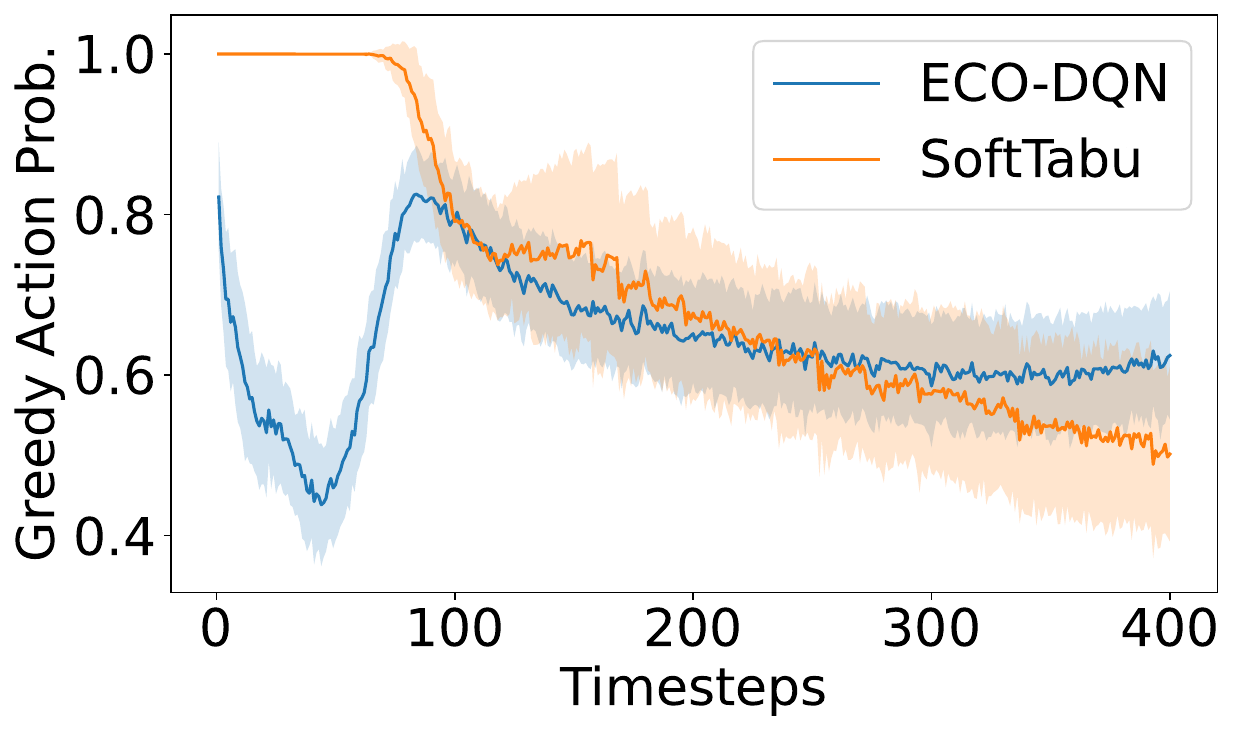}
         \caption{ER200 agent}
         % \label{fig:y equals x}
     \end{subfigure}
     % \hfill
     \begin{subfigure}[b]{0.23\textwidth}
        
         \centering
         \includegraphics[width=\textwidth]{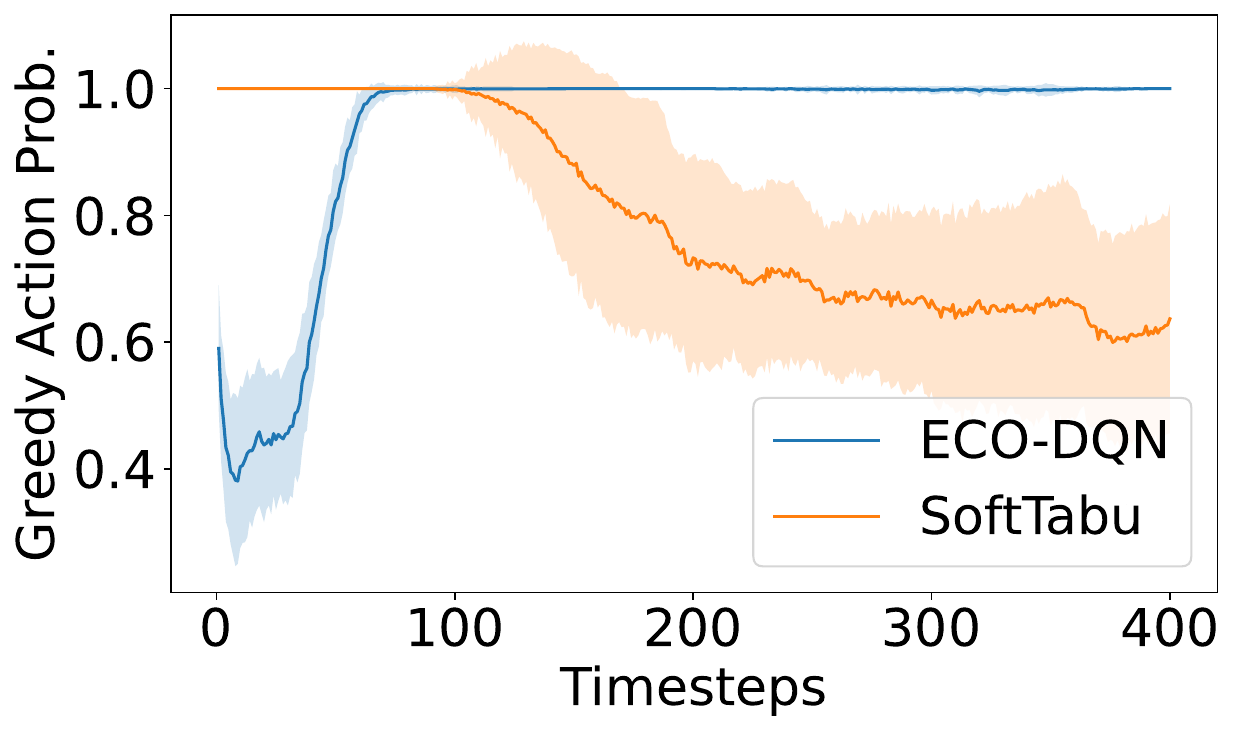}
         % \caption{$y=3\sin x$}
         \caption{BA200 agent}
         % \label{fig:three sin x}
     \end{subfigure}
      \begin{subfigure}[b]{0.23\textwidth}
         
         \centering
         \includegraphics[width=\textwidth]{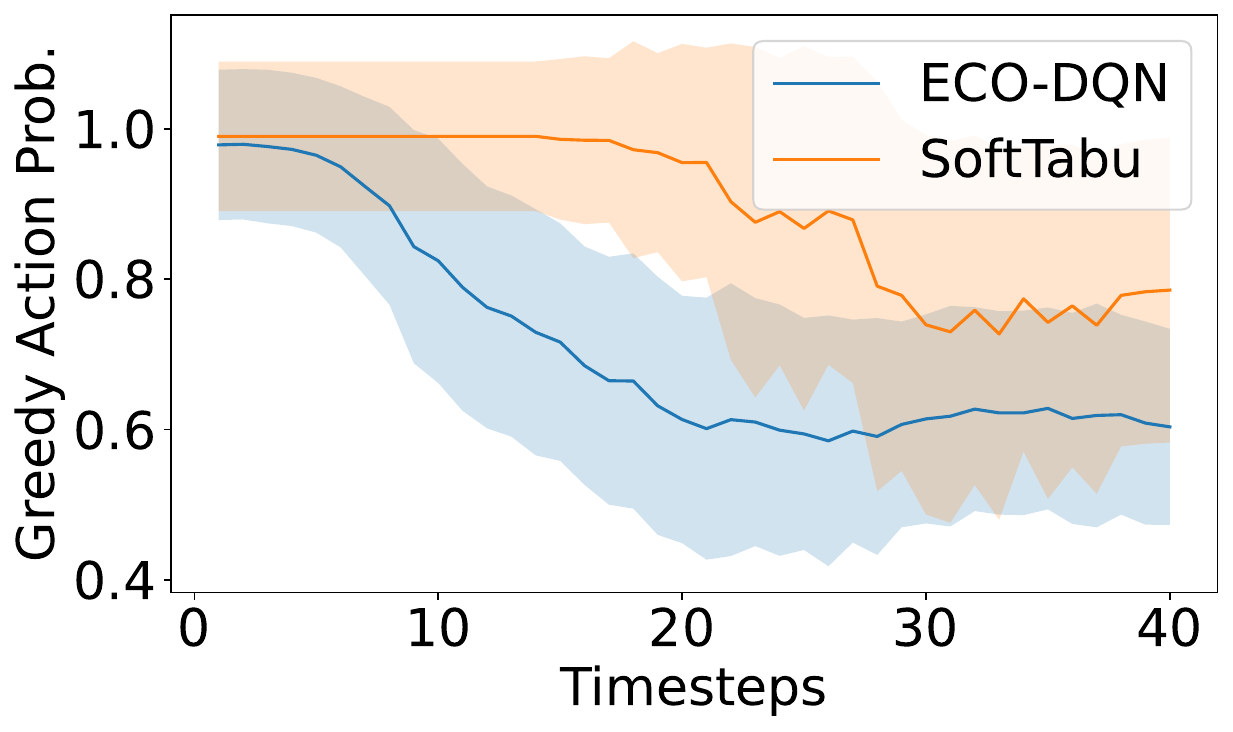}
         \caption{ER20 agent}
         % \label{fig:y equals x}
     \end{subfigure}
     % \hfill
     \begin{subfigure}[b]{0.23\textwidth}
        
         \centering
         \includegraphics[width=\textwidth]{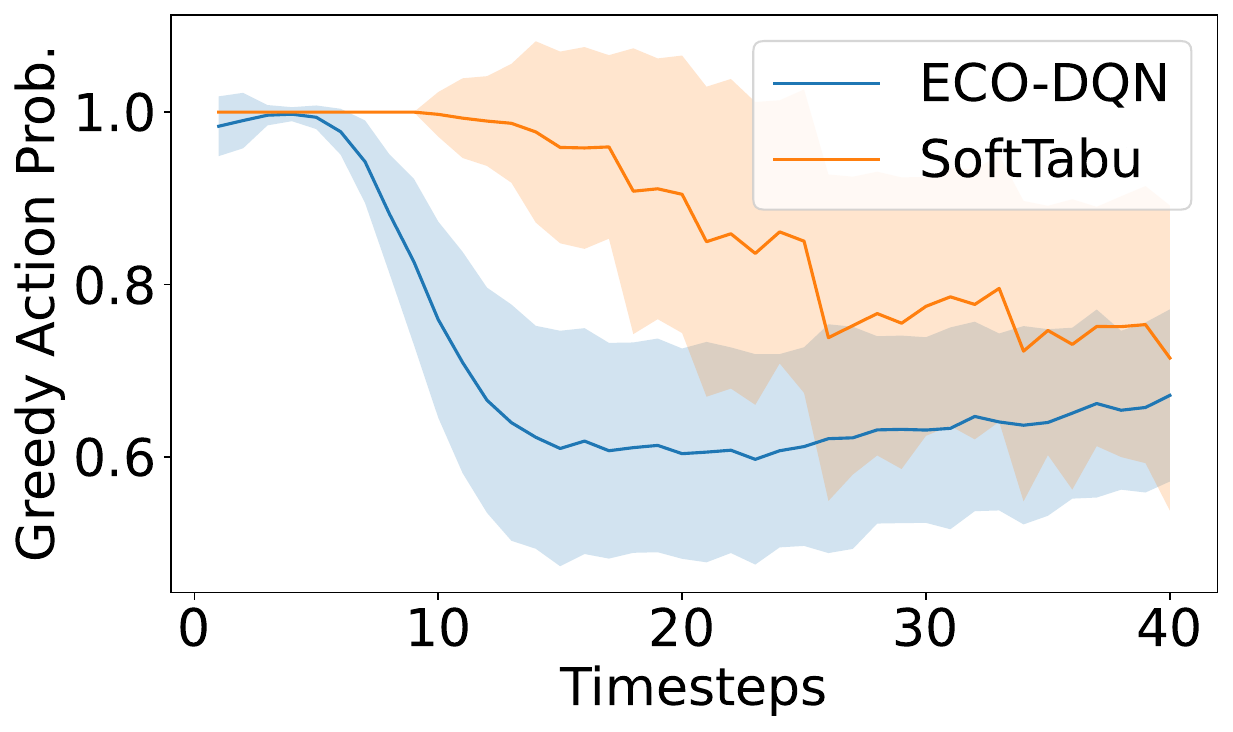}
         % \caption{$y=3\sin x$}
         \caption{BA20 agent}
         % \label{fig:three sin x}
     \end{subfigure}
     \caption{Intra-episode behavior of ECO-DQN and SoftTabu agents averaged across all 100
instances from the validation set for graphs with $|V | =
200$ and $|V | =
20$. }
\label{fig:Intra-episode behaviorMain}
     % \vspace{1em}
        
\end{figure}

Moreover, we note that there is no discernible difference in the performance of ECO-DQN and SoftTabu. This can imply three things: \romannumeral 1) The problem instances are relatively simple, making it difficult for the added features unrelated to TS and GNN used in ECO-DQN to provide significant advantages over SoftTabu. \romannumeral 2) The additional features and GNN utilized in ECO-DQN may not be the primary factors contributing to its superior performance or \romannumeral 3) ECO-DQN and SoftTabu learn similar heuristics for solving the problem, leading to comparable performance despite differences in their architectures and approaches. This is proven to be incorrect. The probability of taking greedy actions at any timestamp greatly varies between the two approaches, as demonstrated in Figure \ref{fig:Intra-episode behaviorMain}.

\subsection{Generalisation to Unseen Graph Types}
\begin{figure}[h]

     \centering
     
     \begin{subfigure}[b]{0.23\textwidth}
         
         \centering
         \includegraphics[width=\textwidth]{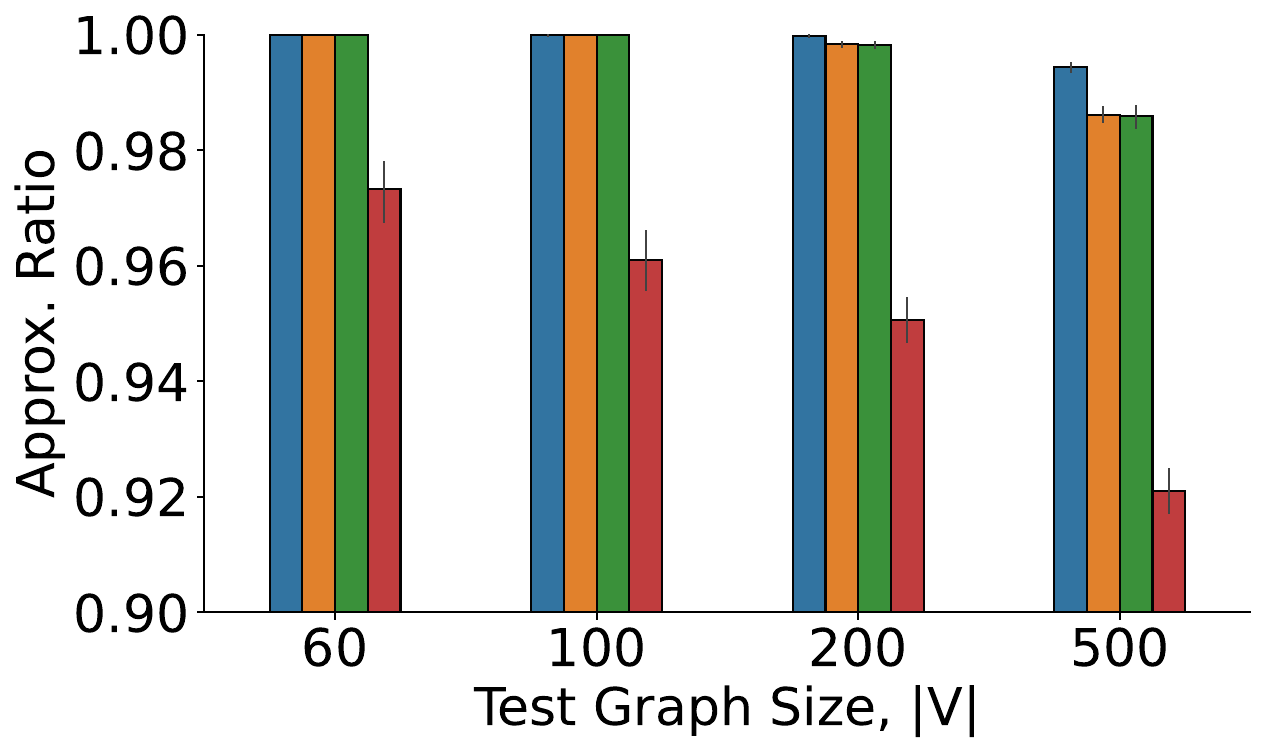}
         \caption{ER(train),ER(test)}
         % \label{fig:y equals x}
     \end{subfigure}
     % \hfill
     \begin{subfigure}[b]{0.23\textwidth}
        
         \centering
         \includegraphics[width=\textwidth]{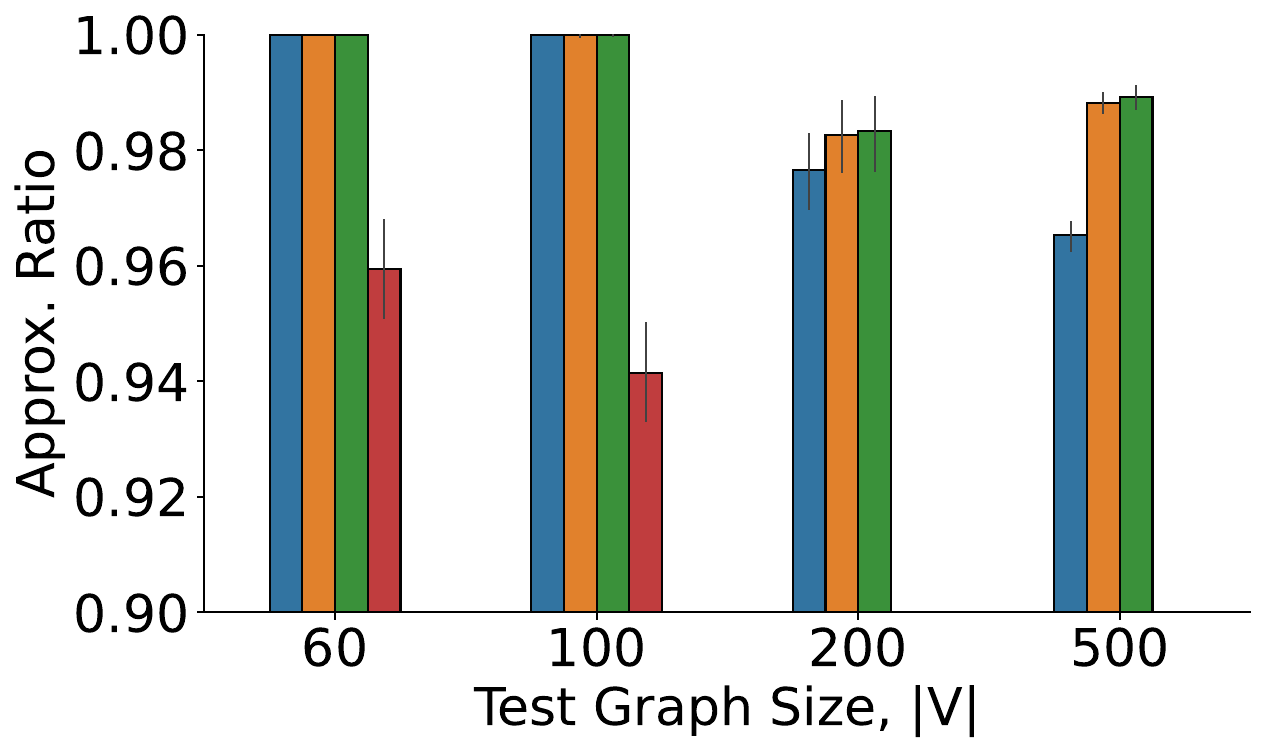}
         \caption{ER(train),BA(test)}
         % \caption{$y=3\sin x$}
         % \label{fig:three sin x}
     \end{subfigure}
     
     % \vspace{1em}

     \begin{subfigure}[b]{0.4\textwidth}
         
         \centering
         \includegraphics[width=\textwidth]{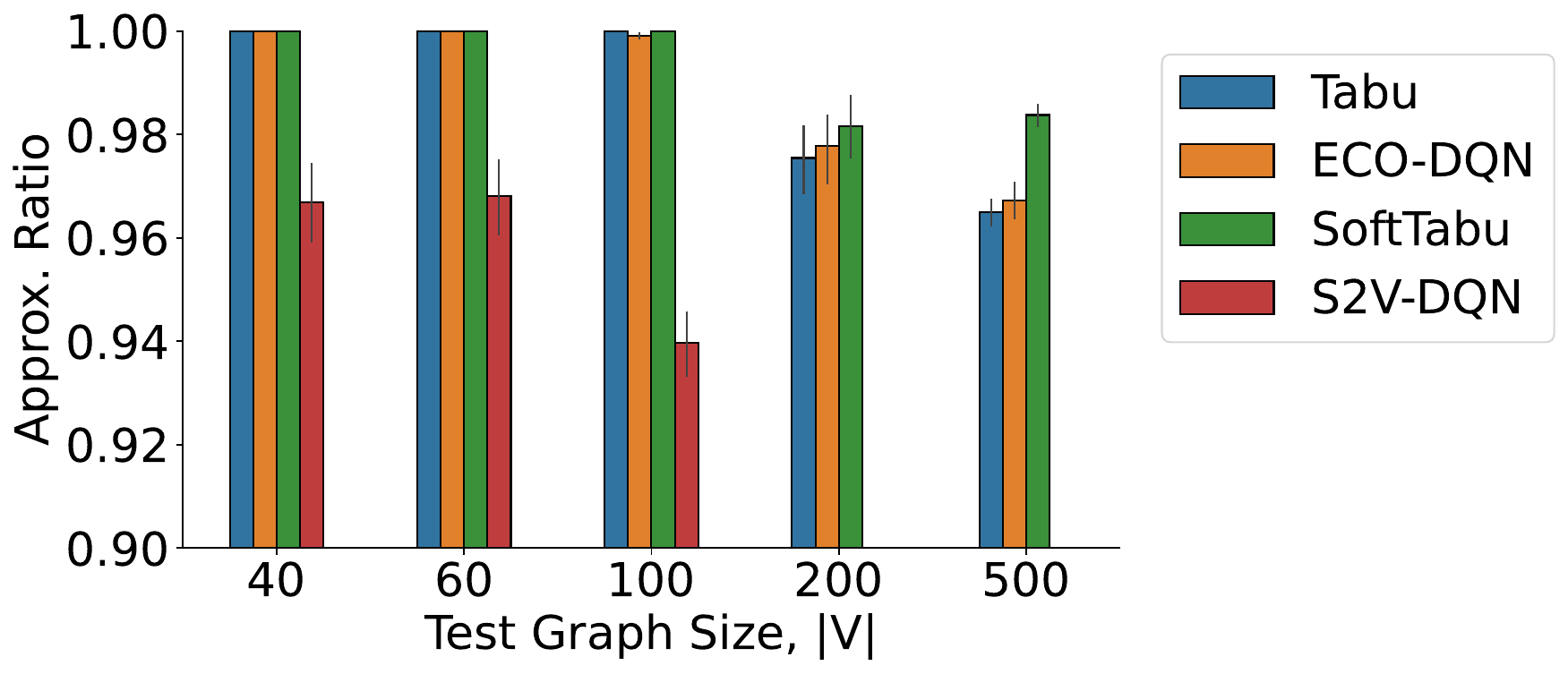}
         \caption{ER(Train),BA(Test)}
         % \label{fig:y equals x}
     \end{subfigure}
    \caption{ Generalization of agents to unseen graph sizes and topologies (a-b) the performance of agents trained on ER and BA
graphs of size $|V |=40$ tested on graphs of up to $|V |=500$ of the same type and (c) a comparison of
how agents trained on ER graphs with $|V|=40$, perform on larger BA graphs.}
\label{fig:GeneralisationMain}
        
\end{figure}

This ability of learned heuristics to perform well on a wide range of distributions, even if these distributions are not represented during training, is a highly desirable characteristic for practical CO problems. ECO-DQN and S2V-DQN exhibited promising performance across a diverse range of graph structures, including those not present in their training data. We run similar experiments for TS and SoftTabu agents to see if they exhibit weaker generalization performance compared to ECO-DQN and S2V-DQN. In our empirical evaluation, we discover that SoftTabu and TS display similar or even superior performance when compared to all learned heuristics, as illustrated in Figure \ref{fig:GeneralisationMain}.
This result is important because it raises the possibility that the generalization of learned heuristics over small instances may not be as novel as the related literature \citep{khalil2017learning,barrett2020exploratory} suggests. If even simple local search heuristics perform well, then the learned policy, which somewhat incorporates these heuristics, may also perform well. 

\subsection{Generalization to Hard Instances and Real World Datasets}
\begin{table*}[h]
\centering
\caption{ Average performance on known benchmarks (best in bold).}
\label{hard-benenchmarksMain}
\begin{tabular}{ccccccc}
\hline
Dataset & Type     & $|V|$ & Tabu  & S2V-DQN & ECO-DQN & SoftTabu \\ \hline
Physics & Regular  & 125   & \textbf{1 }    & 0.928& \textbf{1}       & \textbf{1}       \\
G1-10   & ER       & 800   & 0.989 & 0.950& \textbf{0.990}   & 0.984    \\
G11-G13 & Torodial & 800   & 0.951& 0.919& 0.984   & \textbf{0.988}    \\
G14-G21 & Skew     & 800   & 0.960 & 0.752   & 0.940   & \textbf{0.973}    \\
G22-31  & ER       & 2000  & 0.953 & 0.919   & \textbf{0.981}   & 0.977    \\
G32-34  & Torodial & 2000  & 0.915 & 0.923   & 0.969   & \textbf{0.983}    \\
G35-42  & Skew     & 2000  & 0.949 & 0.694   & 0.864   & \textbf{0.965}    \\ \hline
\end{tabular}
\end{table*}

Our empirical evaluation, up to this point, has been somewhat inconclusive regarding the performance of ECO-DQN. This lack of clarity might be attributed, at least in part, to the lack of harder instances. Specifically, these small instances may be too simple, making it possible for SoftTabu and ECO-DQN to solve them without requiring any special efforts. Consequently, it is important to evaluate the agents on harder instances that possess the ability to differentiate performance levels. To this end, we test agents on the GSET and Physics datasets.
We apply agents that are trained on ER graphs with $|V|=200$, following the experimental setup by \cite{barrett2020exploratory}. They limited their experiments to ER graphs, so we extend the experiments to include all three types of graphs in GSET. Due to the high computational demands, we restrict the number of episodes per graph to 5 for GSET graphs. As these are harder instances, we let agents run for $4|V|$ timesteps to optimize (\cite{barrett2020exploratory} suggested that simply increasing the number of timesteps in an episode increases the average approximation ratio). Notably in Table \ref{hard-benenchmarksMain}, we observe a substantial decline in performance when the ECO-DQN agents are tested on graph distributions other than ER graphs. Even on ER graphs, ECO-DQN demonstrates only marginal improvement. This outcome may be anticipated. When machine learning models are trained on specific datasets, they may learn patterns and heuristics that are tailored to that particular data. However, when presented with unseen or different data (i.e., during testing or deployment), these learned heuristics may not generalize well and could lead to suboptimal performance or poor outcomes.

To investigate further, we generate synthetic skew graphs using the implementation provided in Rudy, a machine-independent graph generator written by G. Rinaldi \citep{ye2003gset}, with descriptions sourced from \citep{helmberg2000spectral}. We train the ECO-DQN on small instances of similar distributions of skew graphs with $200$ vertices, given that ECO-DQN performs worst on skew graphs. This yields an approximate $10\%$ to $15\%$ increase in performance, with the average approximation ratio improving from $\mathbf{0.940}$ to $\mathbf{0.955}$ for $|V|=800$, and from $\mathbf{0.864}$ to $\mathbf{0.945}$ for $|V|=2000$, while still remaining suboptimal compared to SofTabu trained on ER graphs. We provide more details about this experiment in the supplementary material.

\begin{figure}[h]

     \centering
     
     \begin{subfigure}[b]{0.23\textwidth}
         
         \centering
         \includegraphics[width=\textwidth]{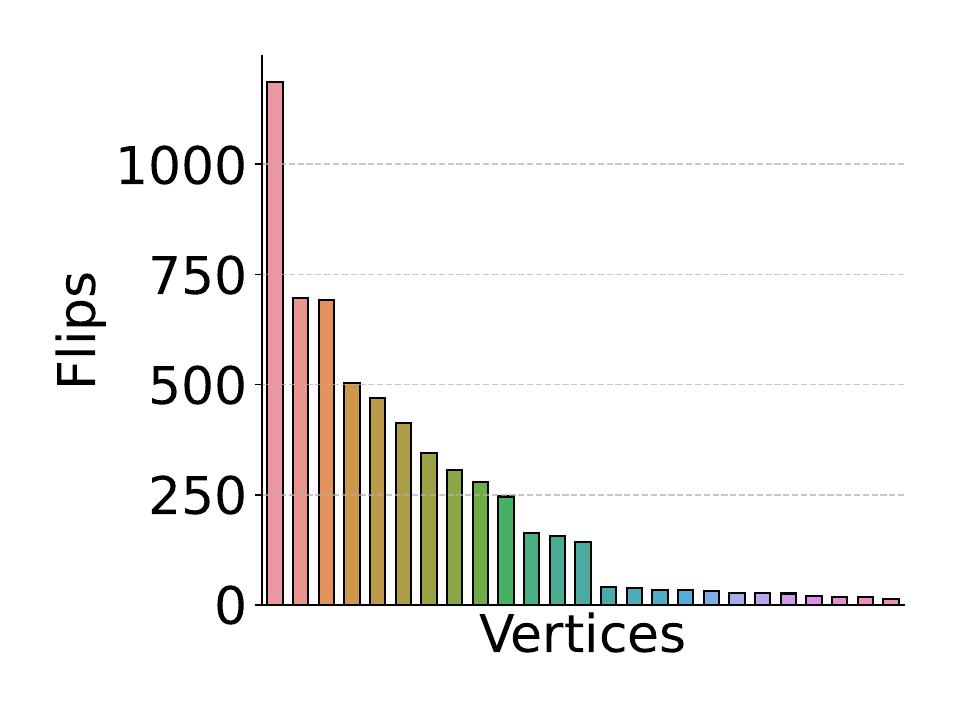}
         \caption{ER(train),ER(test)}
     \end{subfigure}
     % \hfill
     \begin{subfigure}[b]{0.23\textwidth}
        
         \centering
         \includegraphics[width=\textwidth]{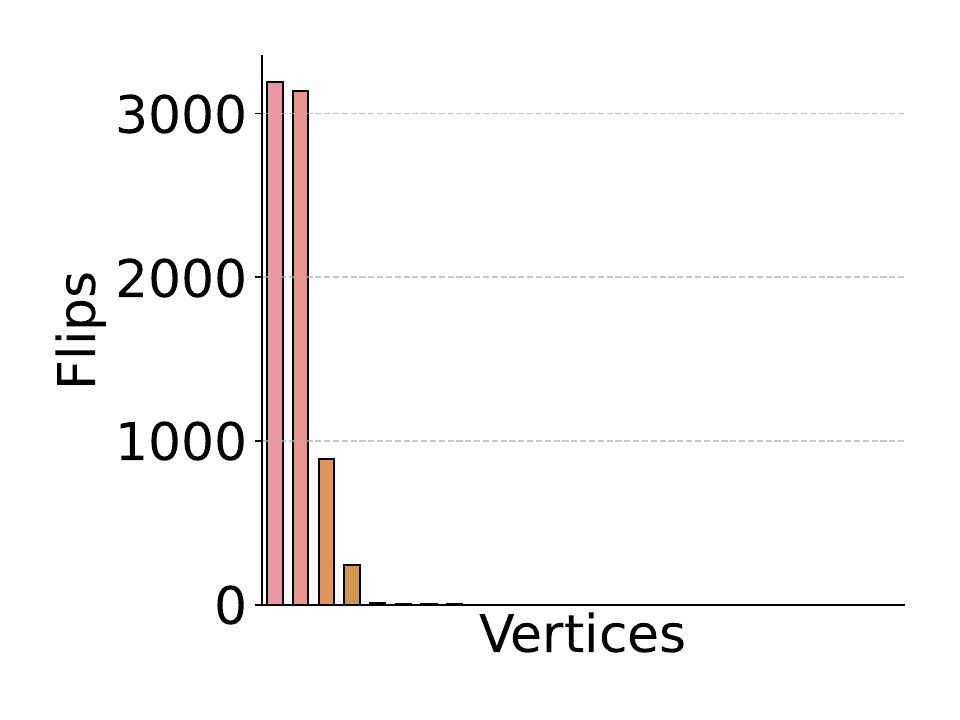}
            \caption{ER(train),Skew(test)}
     \end{subfigure}

     \begin{subfigure}[b]{0.23\textwidth}
         
         \centering
         \includegraphics[width=\textwidth]{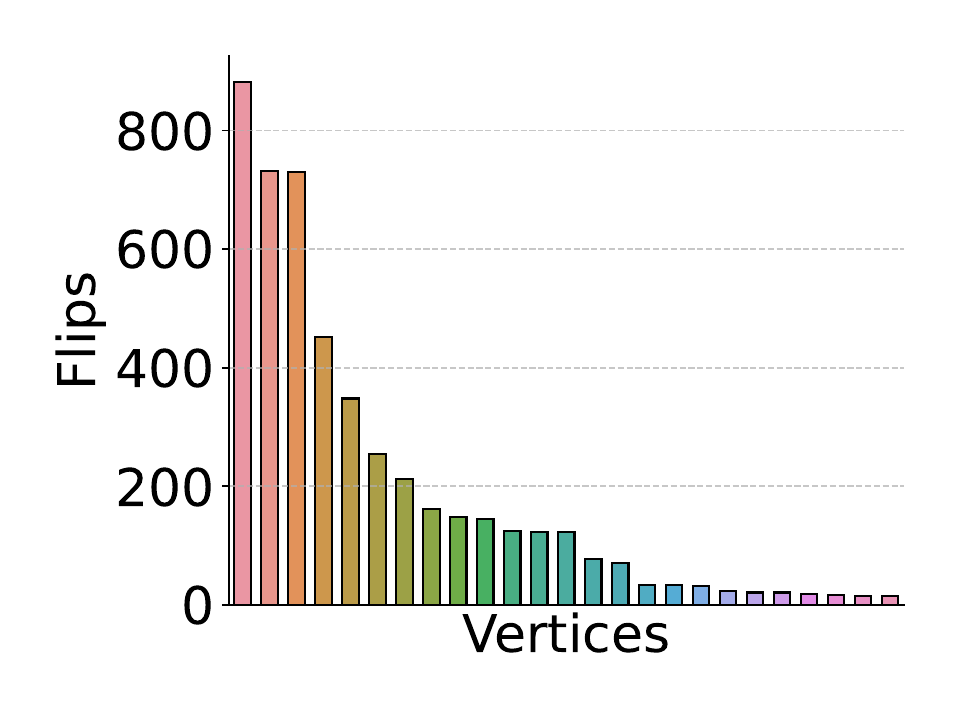}
         \caption{Skew(train),ER(test)}
     \end{subfigure}
     % % \hfill
     \begin{subfigure}[b]{0.23\textwidth}
        
         \centering
         \includegraphics[width=\textwidth]{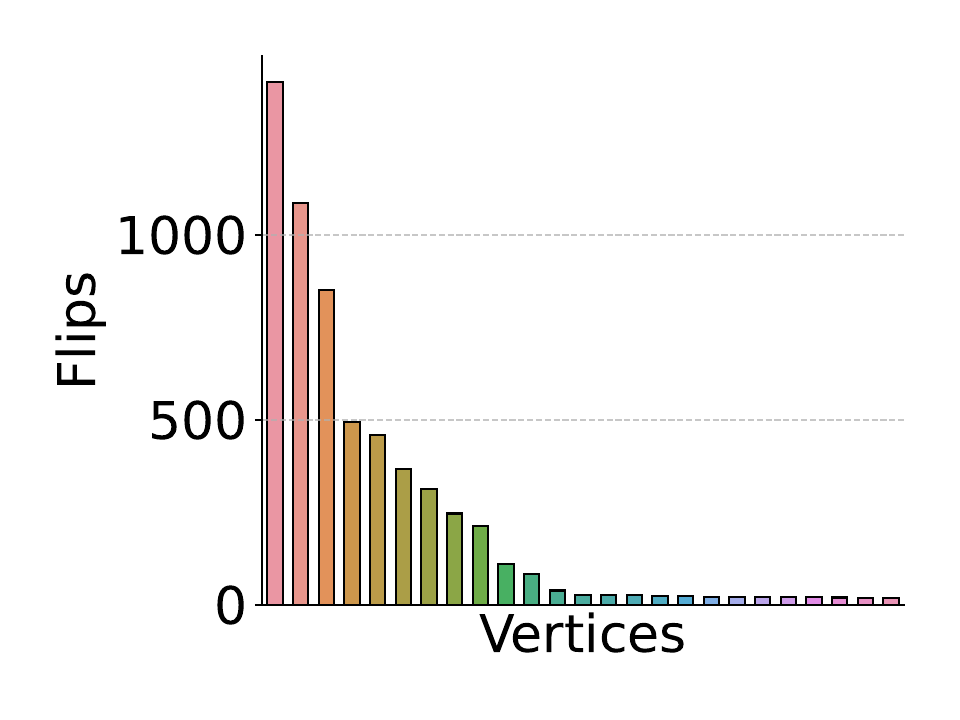}
         \caption{Skew(train),Skew(test)}
     \end{subfigure}
    \caption{Distribution of flips (number of times a vertex state is changed during an episode) of ECO-DQN agents in descending order on a random graph from two distributions with $|V|=2000$ from GSET Dataset, Trained on ER, or Skew Graphs with $|V|=200$. We limit the number of vertices to 25.  }
    \label{fig:action profilingMain}
        
\end{figure}
 
Our experiments indicate that ECO-DQN shows promising generalization to previously unseen graph structures, especially in small instances where simple local search heuristics also perform well. However, when dealing with challenging benchmarks, ECO-DQN may not perform as optimally as simple heuristics or their simple learned counterparts. While it is always possible to find instances where an optimization algorithm seems to perform poorly \citep{wolpert1997no}, our intuition suggests that the reason behind the poor performance of ECO-DQN is that the agent can easily become trapped and cease to explore the search space. Even though the novelty of ECO-DQN lies in the capacity for continuous solution enhancement through exploration, the ECO-DQN agent tends to excessively revisit specific vertices, as shown in Figure \ref{fig:action profilingMain}, thus curbing the exploration of the search space. It becomes clearer when comparing Figure \ref{fig:action profilingMain}b and Figure \ref{fig:action profilingMain}d. When the ECO-DQN agent is trained on skewed graphs, it learns to explore more, resulting in better performance on skewed graphs. We find out that SoftTabu exhibits a more even distribution of flips in an episode (details provided in the supplementary material) and finds better solutions while exploring. This is an interesting outcome in the sense that GNNs seem like a natural choice to solve combinatorial graph problems. However, if the underlying principles of integrated heuristics are something simple, such as exploring to avoid local minima, simpler machine learning models can often adequately learn these principles. Simpler machine learning models can save computational resources and lead to similar or even better results.

\section{EVALUATION for SAT}

\subsection{Problem Formulation}
In this subsection, we formally define the Boolean satisfiability (SAT) problem as follows: Determine whether there exists an assignment \(A\) of truth values to a set of Boolean variables \(V\) such that a given Boolean formula \(F\) in conjunctive normal form evaluates to true: $ F = C_1 \land C_2 \land \ldots \land C_m $ where each clause \(C_i\) is a disjunction of literals: $C_i = l_{i1} \lor l_{i2} \lor \ldots \lor l_{ik_i}$ with literals \(l_{ij}\) being either a variable \(v \in V\) or its negation \(\neg v\).
\subsection{Learned Heuristics}
In this subsection, we delve into GNNSAT \citep{yolcu2019learning} and why we think there is a need for reevaluating GNNSAT.

\paragraph{GNNSAT} In their influential paper, \cite{yolcu2019learning} utilized a GNN to train on factor graph representations of boolean formulas in conjunctive normal form. Their goal was to learn a variable selection heuristic for a stochastic local search algorithm, WalkSAT \citep{selman1993local}. The algorithm demonstrated promising performance compared to WalkSAT. However, it fell short in terms of generalization abilities when compared to the WalkSAT. A direct comparison with SOTA SAT heuristics was omitted. As benchmark instances are extremely large and run time is the metric of performance for comparing with SAT heuristics, the scalability issues of GNNs represent a bottleneck for conducting a fair comparison with SOTA SAT heuristics.

It is important to note that there have been significant improvements \citep{mcallester1997evidence,hoos2002adaptive,li2007combining,mazure1997tabu} in SAT heuristics since the WalkSAT algorithm. However, many of these improvements do not currently represent SOTA SAT heuristics.

While we acknowledge that many neural network-based approaches may face challenges in competing with SOTA SAT solvers in terms of run time due to scalability issues with GNNs, we believe it is still valuable to compare them with algorithms that demonstrate moderate performance. For instance, \cite{kurin2020can} compared their work with MiniSAT\citep{een2003extensible}. Since run time is an issue for GNN, they compared their algorithm with respect to the number of steps (one step is equivalent to choosing a variable and setting it to a value). This comparison provides insights into performance gaps between approaches and showcases the potential of learned SAT solvers to significantly reduce the number of steps. Such reductions, when coupled with advancements in domain-specific processors, can lead to a decrease in runtime, making learned SAT solvers more practical in the future.

Therefore, we focus on evaluating how much improvement GNNSAT provides in terms of steps to solve a problem rather than solely considering absolute runtime. Specifically, our goal is to evaluate whether the performance of the learned heuristic remains competitive in terms of steps, particularly for small instances, when compared to simple heuristics. Addressing this question holds significant implications for the research community in devising strategies to bridge the performance gap. To achieve this objective, we undertake a comparative analysis between SoftTabu and GNNSAT, as both algorithms involve the selection of a variable at each time step and setting its value.

% So it is important to understand how much improvement GNNSAT provides in terms of steps  to solve a problem rather than absolute run time. Specifically, we want to evaluate, for small instances, the performance of the learned heuristic remains competitive (ignoring how much time it takes for a step) when compared to simple heuristics. Addressing this question holds significance for the research community when devising strategies to mitigate the performance gap. To accomplish this objective, we compare SoftTabu with GNNSAT, as both algorithms select a variable at each time-step and set its value.

\subsection{Dataset}
We conduct experiments utilizing randomly generated formulas derived from a diverse family of distributions, encompassing random 3-SAT, clique detection, graph coloring, and dominating set. These specific problem distributions were utilized in the evaluation of GNNSAT.
Random K-SAT problems, especially when critically constrained, are hard, particularly in proximity to the satisfactory/unsatisfactory phase boundary \citep{mitchell1992hard,selman1996generating}.
These problems serve as standard benchmarks for assessing SAT, with the threshold for 3-SAT occurring when problems have approximately $4.26$ times as many clauses as variables. The other three problems fall under the category of NP-complete graph problems. For each of these problems, we randomly sample an Erd{\H{o}}s-R{\'e}nyi graph denoted as $G(N, p)$, where $N$ is the number of vertices and $p$ is the probability of an edge between any two vertices independently from every other edge. We use CNFgen \citep{lauria2017cnfgen} for instance generation and MiniSAT for filtering out unsatisfiable formulas. It is worth noting that we exclude vertex covering—also a benchmark distribution for GNNSAT—due to the unavailability of support for this family of problems in the CNFgen package.

\subsection{Is Scalability the Only Limiting Factor?}

To evaluate the algorithms, we sample $100$ satisfiable formulas from each of the four problem distributions discussed above to obtain evaluation sets. We then run $25$ search trials (with each trial starting at a random initial assignment) for each problem.

\begin{table}[h]
\centering
\caption{Performance of the learned heuristics. In each cell, there are three metrics (top to bottom): the ratio of the average number of steps, the median number of steps, percentage solved following the evaluation methodology of \cite{khudabukhsh2016satenstein}. *Values as reported by \citep{yolcu2019learning} for reference.\newline}
\label{satMain}

\begin{tabular}{cccc}
\hline
  Distribution                                     & SoftTabu &  & GNNSAT* \\ \hline
\multirow{3}{*}{\(\ \text{rand\textsubscript{3}}(50,213)\)}    & 273      &  & 367      \\
                                       & 185      &  & 273      \\
                                       & $96\%$   &  & $84\%$   \\ \hline
\multirow{3}{*}{\(\text{clique\textsubscript{3}}(20,0.05)\)} & 126      &  & 116      \\
                                       & 42       &  & 57       \\
                                       & $100\%$  &  & $100\%$  \\ \hline
\multirow{3}{*}{$\text{color\textsubscript{5}}(20,0.5)$}     & 190      &  & 342      \\
                                       & 77       &  & 223      \\
                                       & $99\%$   &  & $88\%$   \\ \hline
\multirow{3}{*}{\(\text{domset\textsubscript{4}}(12,0.2)\)}  & 76       &  & 205      \\
                                       & 42       &  & 121      \\
                                       & $100\%$  &  & $100\%$  \\ \hline 
\end{tabular}
\end{table}

Since comparing SoftTabu with GNNSAT in runtime will be unfair for the reasons stated above, we follow the evaluation method proposed in \cite{khudabukhsh2016satenstein}. There are three performance metrics for each problem distribution: the average number of steps, the median of the median number of steps (the inner median is over trials on each problem and the outer median is over the problems in the evaluation sets), and the percentage of instances considered solved (the median number of steps less than the allowed number of steps). The main goal of this evaluation is to see if, with the help of GNN, GNNSAT can find the solution in fewer steps than SoftTabu. As shown in Table \ref{satMain}, SoftTabu demonstrates superior performance compared to the learned heuristic, thus shedding light on the performance limitations of the learned local search heuristic. Although GNNSAT is a novel attempt, it seems to show only marginal improvement in performance. Therefore, it is possible that GNNSAT is constrained by both performance and scalability.

%% file: Sections/summary_and_outlook.tex
\section{SUMMARY and OUTLOOK}

% Our empirical evaluation demonstrates that simple heuristics and their learned counterparts can display competitiveness or even demonstrate superior performance. 

Through our empirical evaluations, our goal is to promote an insightful comparison within the research focusing on the intersection of combinatorial optimization and machine learning. In order to provide the research community with valuable guidance, we believe it is imperative to communicate both the strengths and weaknesses of the proposed approaches. Poor instances and baseline selection may give the wrong impression about the performance of learned heuristics. Specifically, it is important to articulate the degree of improvement achieved through integrating classical heuristics with deep learning architectures and conduct a thorough comparison with classical heuristics. This will aid in elucidating the degree to which deep learning architectures enhance integrated heuristics. It assists in ascertaining whether the integration endeavor is justified and warrants the allocation of computational resources, time, and investment necessary for integrating deep learning with classical heuristics.